\documentclass[hidelinks,10pt,conference,letterpaper]{IEEEtran}

\IEEEoverridecommandlockouts
\let\labelindent\relax
\usepackage[table,xcdraw]{xcolor}
\usepackage{subfig}
\usepackage{xspace}
\usepackage{graphicx}
\usepackage{ntheorem}
\usepackage{times,amsmath,epsfig,amssymb,graphicx,amsfonts}
\usepackage{empheq}
\usepackage{subfig}
\usepackage{graphicx}
\usepackage{psfrag}
\usepackage{fge}
\usepackage{amsmath}
\usepackage{setspace}
\usepackage{color}
\usepackage{amsfonts}   
\usepackage{amssymb}    
\usepackage{mathrsfs}
\usepackage{setspace}
\usepackage{dblfloatfix}
\usepackage{tikz}
\usepackage{tkz-tab}

\usepackage{pgfplotstable}
\usepackage{pgfplots}
\usepackage{algorithm}
\usepackage{algcompatible}
\usepackage{lipsum}
\usepackage{color}
\usepackage{comment}
\usepackage{mathrsfs}
\usepackage{epstopdf}
\usepackage{enumerate}
\usepackage{enumitem}
\usepackage{url}
\usepackage{cite}
\usepackage{tabularx}
\usepackage[pass]{geometry}
\usepackage{chngcntr}

\newcommand{\bdmath}{\begin{dmath}}
\newcommand{\edmath}{\end{dmath}}
\newcommand{\beq}{\begin{equation}}
\newcommand{\eeq}{\end{equation}}
\newcommand{\bdm}{\begin{displaymath}}
\newcommand{\edm}{\end{displaymath}}
\newcommand{\bea}{\begin{eqnarray}}
\newcommand{\eea}{\end{eqnarray}}
\newcommand{\beal}{\beq \begin{array}{lll}}
\newcommand{\eeal}{\end{array} \eeq}
\newcommand{\beas}{\begin{eqnarray*}}
\newcommand{\eeas}{\end{eqnarray*}}
\newcommand{\ba}{\begin{array}}
\newcommand{\ea}{\end{array}}
\newcommand{\bit}{\begin{itemize}}
\newcommand{\eit}{\end{itemize}}
\newcommand{\ben}{\begin{enumerate}}
\newcommand{\een}{\end{enumerate}}

\newcommand{\calA}{{\cal A}}
\newcommand{\calB}{{\cal B}}

\newcommand{\calL}{{\cal L}}

\newcommand{\calP}{{\cal P}}

\newcommand{\calR}{{\cal R}}
\newcommand{\calS}{{\cal S}}

\newcommand{\calY}{{\cal Y}}
\newcommand{\calU}{{\cal U}}
\newcommand{\calV}{{\cal V}}

\newcommand{\setN}{\textsf{N}}

\newcommand{\setP}{\textsf{P}}

\newcommand{\hide}[1]{}

\newcommand{\hiddenText}{{\color{gray} hidden text.}}
\newcommand{\hideWithText}[1]{\hiddenText}

\newcommand{\diag}[1]{\mathrm{diag}\left(#1\right)}
\newcommand{\trace}[1]{\mathrm{trace}\left(#1\right)}

\newcommand{\att}[2]{_{#1|#2}}

\newcommand{\myParagraph}[1]{{\bf #1.}\xspace}

\newcommand{\cp}[2]{u_{#1}(#2)}

\newcommand{\metric}{J}
\newcommand{\attack}{\alpha}

\newcommand{\validated}[2]{{#2}}

\newcommand{\elem}{{{v}}}
\newcommand{\function}{{{f}}}

\floatname{algorithm}{Algorithm}

\floatname{algorithm}{Algorithm}
\newtheorem{mydef}{Definition}

\newtheorem{mytheorem}{Theorem}
\newtheorem{mylemma}{Lemma}

\newtheorem{mycorollary}{Corollary}
\newtheorem{myproposition}{Proposition}

\newtheorem{myproblem}{Problem}

\DeclareMathOperator{\sinc}{sinc}

\newcounter{ale}

\newenvironment{liste}{\begin{itemize}}{\end{itemize}}
\newcommand{\aliste}{\begin{liste} \setcounter{ale}{1}}
\newcommand{\zliste}{\end{liste}}

\newcommand{\scaleMathLine}[2][1]{\resizebox{#1\linewidth}{!}{$\displaystyle{#2}$}}

\title{\bf {\LARGE Resilient Active Information Gathering with Mobile Robots }}
\author{Brent Schlotfeldt,{$^{1}$} Vasileios Tzoumas,{$^{1,2}$} Dinesh Thakur,{$^1$} George J.~Pappas{$^{1}$}
\thanks{$^{1}$The authors are with the Department of Electrical and Systems Engineering, University of Pennsylvania, Philadelphia, PA 19104-6228 USA (email: {\fontsize{8}{8}\selectfont\ttfamily\upshape \{brentsc, vtzoumas, pappasg\}@seas.upenn.edu}).}
\thanks{$^{2}$The author is with the Institute for Data, Systems and Society, Massachusetts Institute of Technology, Cambridge, MA 02139 USA (email: {\fontsize{8}{8}\selectfont\ttfamily\upshape vtzoumas@mit.edu}).}
\thanks{This research is partially supported by ARL CRA DCIST W911NF-17-2-0181 and the Rockefeller Foundation.
}
}

\begin{document}
\maketitle

\begin{abstract}
Applications in robotics, such as multi-robot target tracking, involve the execution of information acquisition tasks by teams of mobile robots.  However, in failure-prone or adversarial environments, robots get attacked, their communication channels get jammed, and their sensors fail, resulting in~the withdrawal of robots from the collective task, and, subsequently,~the inability of the remaining active robots to coordinate with each other.  As~a~result, traditional design paradigms become insufficient and, in~contrast, \textit{resilient} designs against system-wide \textit{failures and attacks} become important. In~general, resilient design problems are hard, and even though they often involve objective functions that are monotone and (possibly) submodular, scalable approximation algorithms  for their solution have been hitherto unknown.  In this paper, we provide the first algorithm, enabling the following capabilities:~\textit{minimal communication}, i.e., the algorithm is executed by the robots based only on minimal communication between them;~\textit{system-wide resiliency}, i.e., the algorithm is valid for any number of denial-of-service attacks and failures; and~\textit{provable approximation performance}, i.e., the algorithm ensures for all monotone and (possibly) submodular objective functions a solution that is finitely close to the optimal.  We~support our theoretical analyses with simulated and real-world experiments, by considering an active information acquisition application scenario, namely, \mbox{\textit{multi-robot target tracking}.}
\end{abstract}

\section{Introduction}\label{sec:Intro}
% % % % % % % % % % % % % % % % % % % % % % % % % % % % % % % % % % % % % % % % % % % % % % % % % % % % % % % 

Advances in robotic miniaturization, perception, and  communication~\cite{michini2014robotic,nieto2013multi,kumar2012opportunities,karaman2012high,cadena2016past,8206030,8255576} envision the deployment of robots to support critical missions 
such as: 
\begin{itemize}
\item \textit{Hazardous environmental monitoring}: Deploy a team of mobile robots \textit{to monitor} the radiation flow around a nuclear reactor after an explosion;~\cite{michini2014robotic}
\item \textit{Adversarial-target tracking}: Deploy a team of agile robots \textit{to track} an adversarial target that moves in a cluttered urban environment, aiming to escape;~\cite{nieto2013multi}
\item \textit{Search and rescue}: Deploy a team of  aerial micro-robots \textit{to~localize} people trapped in a burning building;~\cite{kumar2012opportunities}
\end{itemize}
Each of the above scenarios requires the deployment of a mobile team of robots, where each robot needs to be agile; coordinate its motion with its team in a decentralized way; and navigate itself in unknown, complex, and GPS-denied environments, with the objective of gathering the most information about a process of interest. In particular, the problem of designing the motion of a team of mobile robots to infer the state of a process is known as \textit{active information gathering}.

But in all above mission scenarios the robots operate in failure-prone and adversarial environments, where the robots' can get attacked; their communications channels can get jammed; or their sensors can fail.
Therefore, in such failure-prone or adversarial scenarios, 
\textit{resilient} designs against \textit{worst-case} and \textit{system-wide failures and attacks} become important. 

In this paper we formalize for the first time a problem  of \textit{resilient active information gathering} with mobile robots, that goes beyond the traditional objective of (non-resilient) active information gathering, and guards against worst-case failures or attacks that can cause not only the withdrawal of robots from the information gathering task, but also the inability of the remaining robots to jointly optimize their control inputs, due to disruptions to their communication network.

Evidently, resilient active information gathering with mobile robots is a computationally challenging task, since it needs to account for all possible removals of robots from the joint motion-design task, which is a problem of combinatorial complexity.
In particular, this computational challenge motivates one of the primary goals in this paper, namely, to provide a scalable and provably near-optimal \textit{approximation} algorithm for resilient active information gathering with mobile robots.

\myParagraph{Related work} 
Related work on problems of information gathering focuses on  the deployment of either static sensors~\cite{krause2008optimizing, williams2007information, tzoumas2016near} or mobile sensors (mounted on robots)~\cite{hoffmann2010mobile, julian2012distributed, dames2012decentralized, dames2015autonomous, charrow2014approximate,chung2006decentralized,kreucher2005information,atanasov2014information,schlotfeldt2018anytime,atanasov2015decentralized} to monitor a target process. Among these works,
the line of work~\cite{hoffmann2010mobile, julian2012distributed, dames2012decentralized, dames2015autonomous, charrow2014approximate,chung2006decentralized,kreucher2005information,atanasov2014information,atanasov2015decentralized, schlotfeldt2018anytime} is the most relevant to ours, as it considers mobile sensors.  In particular,~\cite{hoffmann2010mobile, julian2012distributed, dames2012decentralized, dames2015autonomous, charrow2014approximate} focus on information gathering tasks over non-Gaussian processes, whereas the remaining~\cite{chung2006decentralized,kreucher2005information,atanasov2014information,atanasov2015decentralized, schlotfeldt2018anytime} focus on information gathering tasks over Gaussian processes.  The advantage in the latter case is that open-loop robot-motion designs are optimal~\cite{atanasov2014information},
an observation that led~\cite{atanasov2014information, atanasov2015decentralized, schlotfeldt2018anytime} to provide the first scalable, non-myopic robot-motion algorithms for active information gathering, along with sub-optimality guarantees. However, in all of these works, there is no resilience to failures or attacks.

In contrast to robotic control, resilient optimization problems have recently received attention in the literature of set function optimization~\cite{orlin2016robust,tzoumas2017resilient,tzoumas2018resilientSequential}.  However,~\cite{orlin2016robust,tzoumas2017resilient,tzoumas2018resilientSequential} focus on the resilient selection of a \textit{small subset} of elements in the event of attacks or failures, whereas the information acquisition problem requires the selection of controls for \textit{all} robots over a {time horizon}. In this paper, we capitalize on the recent results in~\cite{tzoumas2017resilient,tzoumas2018resilientSequential} and seek to bridge the gap between developments in set function optimization and robotic control design to enable critical missions necessitating resilient active information gathering with mobile robots.

\myParagraph{Contributions} 
We make the following contributions:
\begin{itemize}
\item (\textit{Problem definition}) We formalize the problem of \textit{resilient active information gathering with mobile robots} against multi-robot denial-of-service attacks or failures. This is the first work to formalize, address, and demonstrate the importance of this problem. 
\item (\textit{Solution}) We develop the first algorithm for resilient active information gathering with the following properties:
\begin{itemize}
\item \textit{minimal communication}:  it terminates within the same order of communication rounds as state-of-the-art algorithms for (non-resilient) information gathering;
\item \textit{system-wide resiliency}: it is valid for any number of denial-of-service attacks or failures;
\item \textit{provable approximation performance}: for all monotone and (possibly) submodular information gathering objective functions 
in the active robot set (non-failed robots), it ensures a solution close to the optimal.
\end{itemize}
\item (\textit{Simulations}) We conduct simulations in a variety of multi-robot multi-target tracking scenarios, varying the number of robots, targets, and failures. Our simulations validate the benefits of our approach to achieve resilient robotic control against failures or attacks.
\item (\textit{Experiments}) We conduct hardware experiments of multiple quad-rotors tracking static ground targets, to demonstrate visually the necessity for resilient robot motion design against robotic failures or denial-of-service attacks. 
\end{itemize}

\medskip

\myParagraph{Notation}  
Calligraphic fonts denote sets (e.g., $\calA$).  Given a set $\calA$, then $|\calA|$ denotes $\calA$'s cardinality; given also a set $\calB$, then $\calA\setminus\calB$ denotes the set of elements in $\calA$ that are not in~$\calB$. Given a random variable $v$, with mean~$\mu$ and covariance~$\Sigma$, then $v\sim \setN(\mu,\Sigma)$ denotes that~$v$ is a Gaussian random variable.

\section{Problem Statement}\label{sec:prob_form}

We formalize the problem of resilient active information gathering.  To this end, we start with some basic definitions.

\subsection{Basic definitions} 
\label{sec:definitions}

We introduce standard models for the notions \textit{robots}, \textit{target}, \textit{sensors}, and \textit{information objective function}
~\cite{atanasov2014information}.

\setcounter{paragraph}{0}

\paragraph{Robots}
Active information gathering utilizes a team of mobile robots to track the evolution of a target process. We~denote the set of available robots as~$\calV$, 
and model each robot's dynamics as a discrete-time non-linear system:
\begin{align}\label{eq:robot}
x_{i,t} = f_i(x_{i,t-1},u_{i,t-1}),~~~i \in \calV,~~~t=1,2,\ldots,
\end{align}
where the vector $x_{i,t} \in \mathbb{R}^{n_{x_{i,t}}}$ represents the state of robot~$i$ at time $t$, and the vector $u_{i,t} \in \calU_{i,t}$ represents the control input, where $\calU_{i,t}$ is a \textit{finite} set of admissible control inputs.

\paragraph{Target}  
The objective of active information gathering is to track the evolution of a target process.  We model the target's evolution as a standard discrete-time (possibly time-varying) linear system with additive process noise:
\begin{align}\label{eq:target}
\begin{split}
y_{t} &= A_{t-1} y_{t-1} + w_{t-1},~~~t=1,2,\ldots,\\
\end{split}
\end{align}
where the vector $y_t \in \mathbb{R}^{n_{y_t}}$ represents the state of the target at time~$t$, the vector $w_{t-1} \in \mathbb{R}^{n_{y_{t}}}$ represents process noise, and the matrix $A_{t-1}$ has suitable dimension. In addition, we let~$y_0$ be a random variable with covariance $\Sigma\att{0}{0}$, and $w_{t-1}$ be a random variable with zero mean and covariance $W_{t-1}$ such that $w_{t-1}$ is independent of $y_0$ and of $w_{t'-1}$, for all $t'\neq t$.

\paragraph{Sensor measurements}
We consider the sensor measurements to be linearly dependent on the state of the target,\footnote{This standard modeling consideration is without loss of generality whenever linearization over the current estimate of the target's state is possible.} and non-linearly dependent on the robots' state, as follows:
\begin{align}\label{eq:sensors}
\begin{split}
&z_{i,t}=H_{i,t}(x_{i,t})y_t+v_{i,t}(x_{i,t}),~~~i\in\calV,~~~t=1,2,\ldots,
\end{split}
\end{align}
where the vector $z_{i,t} \in \mathbb{R}^{n_{z_{i,t}}}$ is the measurement obtained  at time $t$ by the on-board sensor at robot $i$, the vector  $v_{i,t}(x_{i,t})\in \mathbb{R}^{n_{z_{i,t}}}$ represents measurement noise, and the matrix $H_{i,t}(x_{i,t})$ has suitable dimension. In addition, we let $v_{i,t}(x_{i,t})$ be a random variable with zero mean and covariance $v_{i,t}(x_{i,t})$ such that $v_{i,t}(x_{i,t})$ is independent of $y_0$, of $w_{t'-1}$, and of $v_{i',t'}$ for all $t'\neq t$, and $i' \neq i$.

\paragraph{Information objective function} 
The problem of active information gathering requires the team of robots in $\calV$ to select their control inputs
to maximize the team's tracking capability of a target.  To the latter end, we assume the robots to use a Kalman filtering algorithm to track the evolution of the target over an observation time-horizon $T$\!. Moreover, we consider the robots' collective tracking capability to be quantified by an information objective function, denoted henceforth by~$\metric$, that depends \textit{solely} on the Kalman filter's error covariances across all times $t=1,2,\ldots,T$\!.
Naturally, the Kalman filter's error covariances depend on the robots' control inputs, as well as on both the target process's initial condition $y_0$ and the robots' initial conditions $\{x_{i,0}:i\in \calV\}$. Overall, given an observation time-horizon $T$, it is:
 \begin{align}\label{eq:metric}
 \begin{split}
 \metric=&\metric(u_{1:T}(\calV))\triangleq \\
 &\quad\metric[\Sigma_1(u_{1}(\calV)),\Sigma_2(u_{1:2}(\calV)),\ldots,\Sigma_T(u_{1:T}(\calV))],
 \end{split}
 \end{align}
where $\Sigma_t(u_{1:t}(\calV))$ denotes that Kalman filter's error covariance at time $t$ given the robots' control inputs up to time~$t$, namely, given $\cp{1:t}{\calV}\triangleq \{u_{i,t'}:~u_{i,t'}\in \calU_{i,t'},~i\in\calV,~t' =1,2,\ldots,t\}$.
Examples of information objective functions of the same form as in eq.~\eqref{eq:metric} are the average minimum mean square error $1/T\sum_{t=1}^{T} \trace{\Sigma_t}$, the average confidence-ellipsoid volume $1/T\sum_{t=1}^{T} \log \det (\Sigma_t)$~\cite[Appendix~E]{bertsekas2005dynamic}, as well as information theoretic objectives such as the mutual information $I(y_t | z_{1:t})$ and conditional entropy $h(y_t | z_{1:t})$~\cite{atanasov2014information}, where $z_{1:t}\triangleq \{z_{i,t'}:~i\in \calV,~~t'=1,2,\ldots,t\}$, i.e., $z_{1:t}$ is the set of measurements collected by all robots' across all times.

\subsection{Resilient Active Information Gathering} 
\label{sec:problem_definition}

We define next the main problem in this paper.

\begin{myproblem}
[Resilient Active Information Gathering]
\label{pr:resil_active_inf_acq}
Given a time horizon~$T$, consider a set of robots $\calV$, with dynamics per eq.~\eqref{eq:robot}, with sensing capabilities per eq.~\eqref{eq:sensors}, and with a connected communication network; in addition, consider a target process per eq.~\eqref{eq:target}; moreover, consider an information gathering objective function $\metric$ per eq.~\eqref{eq:metric}; finally, consider a number $\attack \leq |\calV|$.
For all robots $i\in\calV$, and for all times $t=1,2,\ldots,T$\!, find control inputs $u_{i,t}$ to maximize the objective function $\metric$ against a \emph{worst-case} failure or attack to the robots in $\calV$ that causes the removal $\attack$ robots from $\calV$ at the beginning of time ($t=0$), as well as the disruption of all communications among the remaining robots in $\calV$ across all times ($t=1,2,\ldots, T$). 
Formally:
\begin{align}\label{preq:resil_active_inf_acq}
\begin{split}
&\hspace*{-17mm}\max_{\scriptsize\begin{array}{c}
u_{i,t} \in \mathcal{U}_{i,t},i\in\calV,\\
 t=1,2\ldots,T
\end{array}}\min_{\calA\subseteq \calV} \;\;\;\metric(\cp{1:T}{\calV\setminus \calA}):\\
&\hspace*{-3mm}\text{such that, for all } i\in\calV,~~~t=1,2,\ldots,T:\\
&\;y_{t} = A_{t-1} y_{t-1} + w_{t-1},\\
&\;x_{i,t} = f_i(x_{i,t-1},u_{i,t-1}),\\
&\;z_{i,t}=H_{i,t}(x_{i,t})y_{i,t}+v_{i,t}(x_{i,t}),\\
&\;u_{i,t}=u_{i,t}(z_{i,1}, z_{i,2}, \ldots, z_{i,t}),\\
&\;|\calA|\leq \attack,
\end{split}
\end{align}
where for any robot set $\calR\subseteq \calV$ and any time horizon $T$, we let $\cp{1:T}{\calR}\triangleq \{u_{i,t}:~~u_{i,t}\in \calU_{i,t},~~i\in\calR,~~t =1,2,\ldots,T\}$.
\end{myproblem}

We henceforth denote the problem in eq.~\eqref{preq:resil_active_inf_acq} by:
\begin{equation}\label{def:resil_active_inf_acq}
\setP(\calV,\attack),
\end{equation}
where we stress the dependence of the problem only on the set of robots $\calV$, and the maximum number of failures or attacks~$\alpha$. Given an instance of Problem~\ref{pr:resil_active_inf_acq}, and the notation in eq.~\eqref{def:resil_active_inf_acq}, then the (non-resilient) active information gathering problem is the instance of the problem in eq.~\eqref{preq:resil_active_inf_acq} where $\alpha=0$, namely, $\setP(\calV,0)$.  Hence,
Problem~\ref{pr:resil_active_inf_acq} goes beyond the objective of the active information gathering problem $\setP(\calV,0)$, 
by accounting in the planning process for worst-case failures or attacks that (i)~not only may cause the removal of robots from the information gathering task, 
but also, (ii)~they may prevent the remaining robots from jointly re-planning their motion, e.g., due to the caused disruptions to the robots' communication network after the removal of the attacked or failed robots.

\section{Algorithm for \\ Resilient Active Information gathering}\label{sec:alg}

We present the first scalable algorithm for Problem~\ref{pr:resil_active_inf_acq}, whose pseudo-code is described in Algorithm~\ref{alg:dec_resil_coord_decent}; afterwards, we describe the intuition behind it. 

\subsection{Scalable algorithm for Problem~\ref{pr:resil_active_inf_acq}}\label{subsec:alg_description}

\begin{algorithm}[t]
\caption{Scalable algorithm for Problem~\ref{pr:resil_active_inf_acq}.}
\begin{algorithmic}[1]
\REQUIRE  Time horizon~$T$; set of robots $\calV$; dynamics of robots in $\calV$, per eq.~\eqref{eq:robot}; dynamics of target process, per eq.~\eqref{eq:target}; sensing capabilities of robots in $\calV$, per eq.~\eqref{eq:sensors}; information objective function $\metric$, per eq.~\eqref{eq:metric}; number $\attack \leq |\calV|$, per Problem~\ref{pr:resil_active_inf_acq}, that represents the maximum number of possible robot removals from $\calV$.
\ENSURE Control inputs $u_{i,t}$ for all robots $i\in\calV$, and for all times $t=1,2,\ldots,T$.
\medskip
\STATE{Each robot $i\in\calV$ computes the value of the (non-resilient) active information gathering problem:
\begin{equation}\label{pr:marginal_gains_for_each_robot}
\setP(\{i\},0),
\end{equation}
per the notation in eq.~\eqref{def:resil_active_inf_acq}, and denotes it by $q_i$.}\label{line1:step_1}
% \vspace*{-4mm}
\STATE{All robots in $\calV$ find a~subset $\calL$ of $\attack$ robots among them (that is, $\calL\subseteq\calV$ and $|\calL|=\attack$), such that for all robots $i\in \calL$ and all robots $j \in \calV\setminus\calL$, it is $q_i>q_j$;}\label{line1:step_2}
\STATE{Each robot in $\calL$ adopts the control inputs it computed in Algorithm~\ref{alg:dec_resil_coord_decent}'s line~\ref{line1:step_1} by solving the problem in eq.~\eqref{pr:marginal_gains_for_each_robot}.}\label{line1:step_3}
\STATE{The robots in $\calV\setminus \calL$ compute their control inputs by solving the following active information gathering problem:
\begin{equation}\label{pr:active_info_gathering}
\setP(\calV\setminus\calL,0),
\end{equation}
per the notation we introduced in eq.~\eqref{def:resil_active_inf_acq}.}\label{line1:step_4}
\end{algorithmic} \label{alg:dec_resil_coord_decent}
\end{algorithm}

Algorithm~\ref{alg:dec_resil_coord_decent} is composed of four steps: 

\setcounter{paragraph}{0}
\paragraph{Computation of robots' marginal contributions in the absence of attacks (step~\ref{line1:step_1} of Algorithm~\ref{alg:dec_resil_coord_decent})} Each robot $i\in \calV$ solves the problem of active information gathering in eq.~\eqref{pr:marginal_gains_for_each_robot}, which is an instance of Problem~\ref{pr:resil_active_inf_acq} where no other robot participates in the information gathering task, and where no attacks or failures are possible; algorithms to solve such information gathering problems have been proposed in~\cite{atanasov2014information,atanasov2015decentralized, schlotfeldt2018anytime}.  Overall, each robot $i\in \calV$, by solving the problem in eq.~\eqref{pr:marginal_gains_for_each_robot}, computes its marginal contribution to the information gathering task in Problem~\ref{pr:resil_active_inf_acq} in the absence of any other robot in~$\calV\setminus\{i\}$, and in the absence of any attacks and failures.

\paragraph{Computation of robot set $\calL$ with the $\attack$ largest marginal contributions in the absence of attacks (step~\ref{line1:step_2} of Algorithm~\ref{alg:dec_resil_coord_decent})} The robots in $\calV$ share their marginal contribution to the information gathering task, which they computed in Algorithm~\ref{alg:dec_resil_coord_decent}'s step~\ref{line1:step_1},
and decide which subset~$\calL$ of them composes a set of $\attack$ robots with the $\attack$ largest marginal contributions; this procedure can be executed with minimal communication (at most $2|\calV|$ communication rounds), e.g., by accumulating (through the communication network) to one robot all the marginal contributions $\{q_i:i\in\calV\}$, and, then, by letting this robot to select the set $\calL$, and to communicate it back to the rest of the robots. 

\paragraph{Computation of control inputs of robots in $\calL$ (step~\ref{line1:step_3} of Algorithm~\ref{alg:dec_resil_coord_decent})} The robots in the set $\calL$, per Algorithm~\ref{alg:dec_resil_coord_decent}'s step~\ref{line1:step_2}, adopt the control inputs they computed in Algorithm~\ref{alg:dec_resil_coord_decent}'s step~\ref{line1:step_1} (e.g., using the algorithm in~\cite{atanasov2014information}).

\paragraph{Computation of control inputs of robots in $\calV\setminus \calL$ (step~\ref{line1:step_4} of Algorithm~\ref{alg:dec_resil_coord_decent})} Given the set of robots $\calL$, per Algorithm~\ref{alg:dec_resil_coord_decent}'s line~\ref{line1:step_2}, the remaining robots in $\calV\setminus \calL$ jointly solve the problem of active information gathering in eq.~\eqref{pr:active_info_gathering}, which is an instance of Problem~\ref{pr:resil_active_inf_acq} where the robots in $\calL$ do not participate in the information gathering task, and where any attacks or failures are impossible. 
In particular, the robots in~$\calV\setminus \calL$ can jointly solve the problem in eq.~\eqref{pr:active_info_gathering} with minimal communication (at most $2|\calV|$ communication rounds) using the algorithm \textit{coordinate descent}~\cite[Section~IV]{atanasov2015decentralized}.

\subsection{Intuition behind Algorithm~\ref{alg:dec_resil_coord_decent}}\label{subsec:intuition}
The goal of Problem~\ref{pr:resil_active_inf_acq}  is to ensure the success of an information gathering task despite failures or attacks that cause the removal of $\attack$ robots from the task, and,~consequently, disruptions to the robot's communication network (due to the robots' previous removal), which prevent the remaining robots from jointly re-planning their motion.  In~this~context, Algorithm~\ref{alg:dec_resil_coord_decent} aims to fulfill Problem~\ref{pr:resil_active_inf_acq}'s goal~first by separating the set of robots $\calV$ into two subsets ---the set of robots $\calL$, and the (remaining) set of robots $\calV\setminus \calL$ (Algorithm~\ref{alg:dec_resil_coord_decent}'s line~\ref{line1:step_1} and line~\ref{line1:step_2}),--- and~second by designing the robots' control inputs in each of the two sets (Algorithm~\ref{alg:dec_resil_coord_decent}'s line~\ref{line1:step_3} and line~\ref{line1:step_4}).  In~particular, Algorithm~\ref{alg:dec_resil_coord_decent} aims with set~$\calL$  to capture the worst-case attack or failure to $\attack$ robots among the robots in~$\calV$; equivalently, the set $\calL$ is aimed to act as a ``bait'' to an attacker that selects the \textit{best}~$\attack$ robots in $\calV$ (\textit{best} with respect to the robots' contribution towards attaining the goal of Problem~\ref{pr:resil_active_inf_acq}). However, the problem of selecting the \textit{best}~$\attack$ robots in $\calV$ is a combinatorial problem, and, in general, intractable~\cite{Feige:1998:TLN:285055.285059}. 
Therefore, Algorithm~\ref{alg:dec_resil_coord_decent} aims to approximate the best~$\attack$ robots in $\calV$ by letting the set $\calL$ be the set of $\attack$ robots with the $\attack$ largest marginal contributions, and, then, it assigns to them the corresponding control inputs (Algorithm~\ref{alg:dec_resil_coord_decent}'s line~\ref{line1:step_2} and line~\ref{line1:step_3}).  Afterwards, given the set~$\calL$, Algorithm~\ref{alg:dec_resil_coord_decent} assumes the removal of the robots in $\calL$ from~$\calV$, and coordinates the remaining robots in $\calV\setminus \calL$ to jointly plan their motion using a decentralized active information gathering algorithm, such as the coordinated descent algorithm \mbox{proposed in~\cite[Section~IV]{atanasov2015decentralized} (Algorithm~\ref{alg:dec_resil_coord_decent}'s line~\ref{line1:step_4}).}

\section{Performance Guarantees
}\label{sec:performance}

We quantify Algorithm~\ref{alg:dec_resil_coord_decent}'s performance, by bounding the number of robot communication rounds it requires, as well as, by bounding its approximation performance. To this end, {we use the following two notions of curvature for set functions.}\footnote{We focus on properties of set functions  to quantify Algorithm~\ref{alg:dec_resil_coord_decent}'s approximation performance by analyzing the properties of Problem~\ref{pr:resil_active_inf_acq}'s objective function $J$ as a function of the remaining robot set after the removal of a subset of robots from $\calV$ (due to failures or attacks).}

\subsection{Curvature and total curvature of monotone functions}\label{sec:total_curvature}
 
We present the notions of \emph{curvature} and of \emph{total curvature} for non-decreasing set functions.  We start with the notions of \textit{monotonicity}, and of \textit{submodularity} for set functions.

\begin{mydef}[Monotonicity]\label{def:mon}
Consider any \validated{finite ground}{finite} set~$\mathcal{V}$.  The set function $g:2^\calV\mapsto \mathbb{R}$ is 
\emph{non-decreasing} if and only if for any sets $\mathcal{A}\subseteq \calB \subseteq\calV$,  it~holds~$g(\validated{\mathcal{A}'}{\mathcal{B}})\geq g(\mathcal{A})$.
\end{mydef}

\begin{mydef}[Submodularity~{\cite[Proposition 2.1]{nemhauser78analysis}}]\label{def:sub}
Consider any finite set $\calV$.  The set function $g:2^\calV\mapsto \mathbb{R}$ is \emph{submodular} if and only if
for any sets $\mathcal{A}\subseteq \validated{\mathcal{A}'}{\mathcal{B}}\subseteq\calV$, and any element $\elem\in \calV$, it \validated{is}{holds}  
$g(\mathcal{A}\cup \{\elem\})\!-\!g(\mathcal{A})\geq g(\validated{\mathcal{A}'}{\mathcal{B}}\cup \{\elem\})\!-\!g(\validated{\mathcal{A}'}{\mathcal{B}})$.
\end{mydef}
In words, a set function $g$ is submodular if and only if it satisfies a diminishing returns property where
for any $\mathcal{A}\subseteq \mathcal{V}$ and $\elem\in \mathcal{V}$, the drop $g(\mathcal{A}\cup \{\elem\})-g(\mathcal{A})$ is~non-increasing.

\begin{mydef}\label{def:curvature}
\emph{\textbf{(Curvature of monotone submodular functions~\cite{conforti1984curvature})}}
Consider a finite set $\mathcal{V}$ and a non-decreasing submodular set function $g:2^\mathcal{V}\mapsto\mathbb{R}$ such that (without loss of generality) for any elements $\elem \in \mathcal{V}$, it is  $g(\elem)\neq 0$.  The curvature of $g$ is defined as follows: \begin{equation}\label{eq:curvature}
\kappa_g\triangleq 1-\min_{\elem\in\mathcal{V}}\frac{g(\mathcal{V})-g(\mathcal{V}\setminus\{\elem\})}{g(\elem)}.
\end{equation}
\end{mydef}

Notably, the above notion of curvature implies that for any non-decreasing submodular set function $g$, it is $0 \leq \kappa_g \leq 1$.

\begin{mydef}\label{def:total_curvature}
\emph{\textbf{(Total curvature of non-decreasing functions~\cite[Section~8]{sviridenko2017optimal})}}
Consider a finite set $\mathcal{V}$ and a monotone set function $g:2^\mathcal{V}\mapsto\mathbb{R}$.  The total curvature of $g$ is defined as follows: 
\begin{equation}\label{eq:total_curvature}
c_g\triangleq 1-\min_{v\in\mathcal{V}}\min_{\mathcal{A}, \mathcal{B}\subseteq \mathcal{V}\setminus \{v\}}\frac{g(\{v\}\cup\mathcal{A})-g(\mathcal{A})}{g(\{v\}\cup\mathcal{B})-g(\mathcal{B})}.
\end{equation}
\end{mydef}

The above notion of total curvature implies that for any non-decreasing set function $g$, it is $0 \leq c_g \leq 1$. Moreover, to connect the notion of total curvature with that of curvature, we note that when a function $g$ is non-decreasing and submodular, then the two notions coincide, i.e., $c_g=\kappa_g$.

\subsection{Performance analysis for Algorithm~\ref{alg:dec_resil_coord_decent}}\label{subsec:perf_decentralized_resil_alg}

We quantify Algorithm~\ref{alg:dec_resil_coord_decent}'s approximation performance, as well as, the number of communication rounds it requires.

\begin{mytheorem}\label{th:per_alg_dec_resil_coord_decent}
\emph{\textbf{(Performance of Algorithm~\ref{alg:dec_resil_coord_decent})}}
Consider an instance of Problem~\ref{pr:resil_active_inf_acq},
and the definitions:
\begin{itemize}
\item let the number $\metric^\star$ be the (optimal) value to Problem~\ref{pr:resil_active_inf_acq}, i.e., it is $\metric^\star\triangleq \setP(\calV,\attack)$;
\item given any control inputs $u_{1:T}(\calV)$ for the robots in $\calV$, let the set $\calA^\star[u_{1:T}(\calV)]$ be a (worst-case) removal of $\alpha$ robots from $\calV$, i.e., $\calA^\star[u_{1:T}(\calV)]\triangleq \arg\min_{\calA\subseteq \calV} \metric(\cp{1:T}{\calV\setminus \calA})$;
\item given any removal of a subset of robots $\calA$ from the robot set $\calV$ (due to attacks or failures), call the remaining robot set $\calV\setminus \calA$ \emph{active robot set}.
\end{itemize}
Finally, consider the robots in~$\calV$ to solve optimally the problems in Algorithm~\ref{alg:dec_resil_coord_decent}'s step~\ref{line1:step_1}
and step~\ref{line1:step_4}, using an algorithm that terminates in~$\rho$ communication rounds.

\begin{enumerate}
\item{(Approximation performance)} Algorithm~\ref{alg:dec_resil_coord_decent} returns control inputs $u_{1:T}(\calV)$ such that:
\begin{itemize}
    \item If the objective function $\metric$ is non-decreasing and submodular in the active robot set, and (without loss of generality) $\metric$ is non-negative and $\metric[u_{1:T}(\emptyset)]=0$, then, it is: 
\begin{equation}\label{ineq:bound_sub}
\!\!\frac{\metric\{u_{1:T}[\calV\setminus\calA^\star(u_{1:T}(\calV)]\}}{\metric^\star}\geq \max\left(1-\kappa_\metric,\frac{1}{1+\alpha}\right),%\ell,
    \end{equation}
    where $\kappa_\metric$ is the curvature of $\metric$ (Definition~\ref{def:curvature}).
    
\item If the objective function $\metric$ is non-decreasing in the active robot set, and (without loss of generality) $\metric$ is non-negative and $\metric[u_{1:T}(\emptyset)]=0$, then, it is:
	\begin{equation}\label{ineq:bound_non_sub}
	 \frac{\metric\{u_{1:T}[\calV\setminus\calA^\star(u_{1:T}(\calV)]\}}{\metric^\star}\geq (1-c_\metric)^2\!\!,%\ell,
	\end{equation}
    where $c_\metric$ is the total curvature of $\metric$ (Definition~\ref{def:total_curvature}).

\end{itemize}

\item{(Communication rounds)} Algorithm~\ref{alg:dec_resil_coord_decent} terminates in at most $2|\calV|+\rho$ communication rounds.
\end{enumerate}
\end{mytheorem}

Theorem~\ref{th:per_alg_dec_resil_coord_decent}  implies on Algorithm~\ref{alg:dec_resil_coord_decent}'s performance:
\setcounter{paragraph}{0}
\paragraph{{Near-optimality}} For both  monotone submodular and merely monotone information objective functions, Algorithm~\ref{alg:dec_resil_coord_decent} guarantees a value for Problem~\ref{pr:resil_active_inf_acq} 
which is 
finitely close to the optimal.  
For example,
per ineq.~\eqref{ineq:bound_sub}, 
the~approximation factor of Algorithm~\ref{alg:dec_resil_coord_decent} is bounded by $1/(1+\attack)$, which, for any finite number of robots $|\calV|$, 
is non-zero.

\paragraph{{Approximation difficulty}}
For both monotone submodular and merely monotone information objective functions,  when the curvature $\kappa_\metric$ or the total curvature $c_\metric$, respectively, tend to zero, Algorithm~\ref{alg:dec_resil_coord_decent} becomes exact 
since for $\kappa_\metric\rightarrow 0$ and $c_\metric\rightarrow 0$ the terms $1-\kappa_\metric$ and $1-c_\metric$ in ineq.~\eqref{ineq:bound_sub} and ineq.~\eqref{ineq:bound_non_sub} tend to $1$.
Overall, Algorithm~\ref{alg:dec_resil_coord_decent}'s curvature-dependent
approximation bounds make a first step towards separating
the classes of monotone submodular and merely monotone information objective functions into
functions for which Problem~\ref{pr:resil_active_inf_acq}
can be approximated well (low curvature functions), and functions for which it cannot \mbox{(high curvature functions).}

Overall, Theorem~\ref{th:per_alg_dec_resil_coord_decent} quantifies Algorithm~\ref{alg:dec_resil_coord_decent}'s approximation performance when the robots in~$\calV$ solve optimally the problems in Algorithm~\ref{alg:dec_resil_coord_decent}'s step~\ref{line1:step_1} and step~\ref{line1:step_4}. However, the problems in Algorithm~\ref{alg:dec_resil_coord_decent}'s step~\ref{line1:step_1} and step~\ref{line1:step_4} are computationally challenging, and only approximation algorithms are known for their solution, among which the recently proposed \emph{coordinate descent}~\cite[Section~IV]{atanasov2015decentralized}; in particular, coordinate descent has the advantages of being scalable and of having provable approximation performance.  We next quantify Algorithm~\ref{alg:dec_resil_coord_decent}'s performance when the robots in~$\calV$  solve the problem in Algorithm~\ref{alg:dec_resil_coord_decent}'s step~\ref{line1:step_4} using {coordinate descent} 
(we refer the reader to AppendixA for a description of {coordinate descent}).

\begin{myproposition}\label{prop:alg_per_with_coordinate_descent}
Consider an instance of Problem~\ref{pr:resil_active_inf_acq}, and the notation introduced in Theorem~\ref{th:per_alg_dec_resil_coord_decent}.  Finally, consider the robots in~$\calV$ to solve the problem in Algorithm~\ref{alg:dec_resil_coord_decent}'s step~\ref{line1:step_1} optimally,
and the problem in Algorithm~\ref{alg:dec_resil_coord_decent}'s  step~\ref{line1:step_4} using \emph{coordinate descent}~\cite[Section~IV]{atanasov2015decentralized}.

\begin{enumerate}
\item{(Approximation performance)} Algorithm~\ref{alg:dec_resil_coord_decent} returns control inputs $u_{1:T}(\calV)$ such that:
\begin{itemize}
    \item If the objective function $\metric$ is non-decreasing and submodular in the active robot set, and (without loss of generality) $\metric$ is non-negative and $\metric[u_{1:T}(\emptyset)]=0$, then, it is: 
\begin{equation}\label{ineq:bound_sub_cd}
\frac{\metric(u_{1:T}(\calV))}{\metric^\star}\geq \frac{\max\left(1-\kappa_\metric,1/(1+\alpha)\right)}{2}.
\end{equation}
    
\item If the objective function $\metric$ is non-decreasing in the active robot set, and (without loss of generality) $\metric$ is non-negative and $\metric[u_{1:T}(\emptyset)]=0$, then, it is: 
\begin{equation}\label{ineq:bound_non_sub_cd}
\frac{\metric(u_{1:T}(\calV))}{\metric^\star}\geq \frac{(1-c_\metric)^3}{2}.
\end{equation}
\end{itemize}

\item{(Communication rounds)} Algorithm~\ref{alg:dec_resil_coord_decent} terminates in at most $3|\calV|$ communication rounds.% among the robots in~$\calV$.
\end{enumerate}
\end{myproposition}

Proposition~\ref{prop:alg_per_with_coordinate_descent} implies on Algorithm~\ref{alg:dec_resil_coord_decent}'s  performance:
 \setcounter{paragraph}{0}
\paragraph{Approximation performance for low curvature}
For both monotone submodular and merely monotone information objective functions,  when the curvature $\kappa_\metric$ or the total curvature $c_\metric$, respectively, tend to zero, Algorithm~\ref{alg:dec_resil_coord_decent} recovers the same approximation performance as that of the state-of-the-art algorithms for (non-resilient) active information gathering Algorithm~\ref{alg:dec_resil_coord_decent} calls as subroutines. For example, for submodular information objective functions, the algorithm for active information gathering \textit{coordinate descent}~\cite[Section~IV]{atanasov2015decentralized} has approximation performance at least $1/2$ the optimal~\cite[Theorem~2]{atanasov2015decentralized}, and,  per Proposition~\ref{prop:alg_per_with_coordinate_descent}, when Algorithm~\ref{alg:dec_resil_coord_decent} calls as subroutine this algorithm, it has approximation performance at least $(1-\kappa_J)/2$ the optimal, which tends to $1/2$  for $\kappa_\metric\rightarrow 0$.

\paragraph{Approximation performance for no failures or attacks}
For submodular information objective functions, and for zero number of failures or attacks ($\attack=0$), Algorithm~\ref{alg:dec_resil_coord_decent}'s approximation performance becomes the same as that of the state-of-the-art algorithms for (non-resilient) active information gathering Algorithm~\ref{alg:dec_resil_coord_decent} calls as subroutines.  In particular, for submodular information objective functions, the algorithm for active information gathering \textit{coordinate descent}~\cite[Section~IV]{atanasov2015decentralized} has approximation performance at least $1/2$ the optimal, and,  per Proposition~\ref{prop:alg_per_with_coordinate_descent}, when Algorithm~\ref{alg:dec_resil_coord_decent} calls as subroutine this algorithm, it has approximation performance at least $1/2$ the optimal  for $\attack=0$, since it is  $1/(1+0)=1$ in ineq.~\eqref{ineq:bound_sub_cd}.

\paragraph{Minimal communication}
Algorithm~\ref{alg:dec_resil_coord_decent}, even though it goes beyond the objective of (non-resilient) active information gathering, by accounting for attacks or failures, it terminates within the same order of communication rounds as state-of-the-art algorithms for (non-resilient) active information gathering. In particular, the algorithm for active information gathering \textit{coordinate descent}~\cite[Section~IV]{atanasov2015decentralized} terminates in at most $|\calV|$ rounds, and, per Proposition~\ref{prop:alg_per_with_coordinate_descent}, when Algorithm~\ref{alg:dec_resil_coord_decent} calls as a subroutine this algorithm, then it terminates in at most $3|\calV|$ rounds; \mbox{evidently, $|\calV|$ and $3|\calV|$ have the same order.}

\myParagraph{Summary of theoretical results}
Overall, Algorithm~\ref{alg:dec_resil_coord_decent} is the first algorithm for Problem~\ref{pr:resil_active_inf_acq}, and it enjoys the following:
\begin{itemize}
\item \textit{minimal communication}:  Algorithm~\ref{alg:dec_resil_coord_decent} terminates within the same order of communication rounds as state-of-the-art algorithms for (non-resilient) information gathering;
\item \textit{system-wide resiliency}: Algorithm~\ref{alg:dec_resil_coord_decent} is valid for any number of denial-of-service attacks and failures;
\item \textit{provable approximation performance}: Algorithm~\ref{alg:dec_resil_coord_decent} ensures for all monotone and (possibly) submodular objective functions a solution finitely close to the optimal.
\end{itemize}

\section{Application: \\ Multi-target tracking with mobile robots}\label{sec:simulations_experiments}

We motivate the importance of Problem~\ref{pr:resil_active_inf_acq}, as well as, demonstrate the performance of Algorithm~\ref{alg:dec_resil_coord_decent}, by considering an application of active information gathering, namely, \textit{multi-target tracking with mobile robots}. 
In particular, the application's setting is as follows: a~team~$\mathcal{V}$ of mobile robots is tasked with tracking the position of~$M$ moving targets. In more detail, each robot moves according to unicycle dynamics on $SE(2)$, discretized with a sampling period $\tau$: 
\begin{align*}
\scaleMathLine[.88]{
\begin{pmatrix}
x_{t+1}^1 \\ x_{t+1}^2 \\ \theta_{t+1}
\end{pmatrix} = 
\begin{pmatrix}
x_t^1 \\x_t^2 \\ \theta_t 
\end{pmatrix} + 
\begin{pmatrix}
\nu \sinc(\frac{\omega \tau}{2}) \cos (\theta_t + \frac{\omega \tau}{2})\\
\nu \sinc(\frac{\omega \tau}{2}) \sin (\theta_t + \frac{\omega \tau}{2})\\
\tau \omega
\end{pmatrix}.}
\end{align*}
The set of admissible controls is given by $\mathcal{U} :=$ $\{(\nu,\omega)$ $: $ $\nu \in \{1,3\}$ m/s, $\omega \in \{0, \pm 1, \pm 3\}$ rad/s\}. 

The targets move according to double integrator dynamics, corrupted with additive Gaussian noise.
% , where $ y \in \mathbf{R}^{n_y}$ is the targets' state. 
For~$M$ targets, their state at time $t$ is $y_t = \begin{bmatrix}y_{t,1}^\top y_{t,2}^\top, \ldots, y_{t,M}^\top \end{bmatrix}^\top$
% with $n_y=4M$, 
where $y_{t,m}$ contains the planar coordinates and velocities of the $m$-th target, denoted by ($y^1$, $y^2$, $\dot{y^1}$, $\dot{y^2}$). The~model is:
\begin{align*}
\scaleMathLine[1]{
y_{t+1,m} = A\begin{bmatrix} I_2 &\tau I_2 \\
					0 & I_2  \end{bmatrix}y_{t,m} + w_t, \hspace{3mm} w_t \sim \setN \begin{pmatrix}0,q \begin{bmatrix} \tau^3/3 I_2 & \tau^2/2 I_2 \\ \tau^2/2 I_2 & \tau I_2 \end{bmatrix}\end{pmatrix}.}                    
\end{align*}

The sensor observation model consists of a range and bearing for each target $m \in \{0,\ldots, M-1\}$:
% , where $M$ is the total number of targets in the environment:
\begin{equation*}
\begin{aligned}
&z_{t,m} = h(x_t,y_{t,m}) + v_t, \hspace{3mm} v_t \sim \setN \begin{pmatrix}0, V(x_t,y_{t,m}) \end{pmatrix};\\
&\scaleMathLine[0.88]{h(x,y_m) = \begin{bmatrix}r(x,y_m)\\\alpha(x,y_m)\end{bmatrix} := \begin{bmatrix} \sqrt{(y^1 - x^1)^2 + (y^2 -x^2)^2} \\ \tan^{-1} ((y^2 - x^2)(y^1 - x^1)) - \theta \end{bmatrix}.}
\end{aligned} 
\end{equation*}
We note that since the sensor observation model is non-linear, we linearize it around the predicted target trajectory $y \neq x$:
\begin{align*}
\scaleMathLine[1]{\nabla_y h(x,y_m) = \frac{1}{r(x,y_m)} 
\begin{bmatrix} (y^1 - x^1) & (y^2 - x^2) & 0_{1x2} \\
-\sin( \theta + \alpha(x,y_m)) & \cos (\theta + \alpha(x,y_m)) & 0_{1x2} \\ \end{bmatrix}.}
\end{align*}
The observation model for the joint target state can then be expressed as a block diagonal matrix containing the linearized observation models for each target along the diagonal, i.e.,
\begin{align*}
H \triangleq \diag{\nabla_{y_1}h(x,y_1), \ldots, \nabla_{y_M}h(x,y_M)}.
\end{align*}
The sensor noise covariance grows linearly in range and in bearing, up to $\sigma_r^2$, and $\sigma_b^2$, where $\sigma_r$ and $\sigma_b$ are the standard deviation of the range and the bearing noise, respectively. The model here also includes a limited range and field of view, denoted by the parameters $r_{sense}$ and $\psi$, respectively. \\

\begin{figure}[t]
\centering
\includegraphics[width=0.8\linewidth]{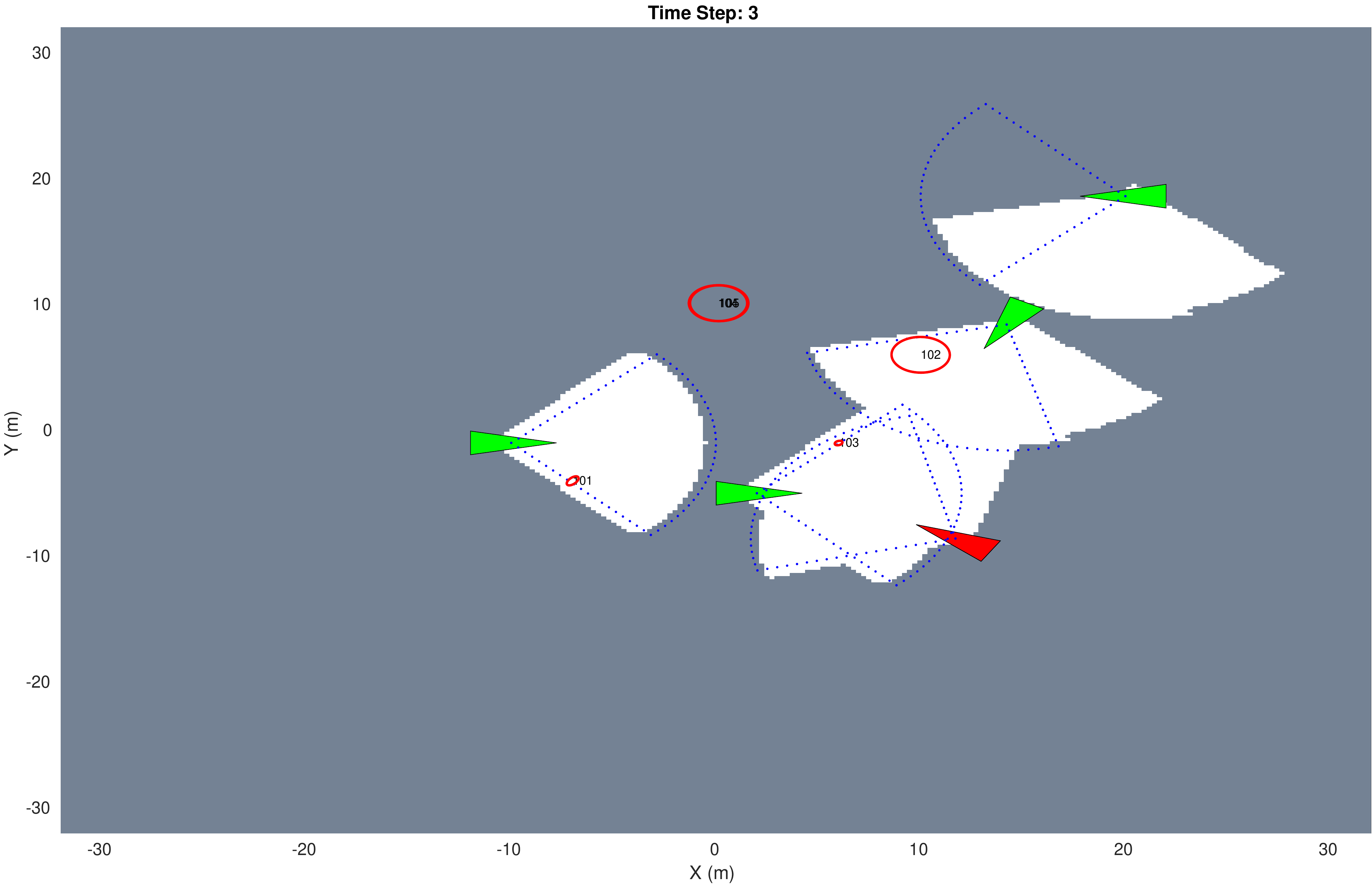}
\caption{\small Simulation environment depicting five robots. The jammed robot is indicated in red.}
  \label{fig:sim_ex}
\end{figure}

\indent Finally, as information objective function, in the simulations we use the average log determinant of the covariance matrix\cite{atanasov2015decentralized,schlotfeldt2018anytime}. Overall, we solve an instance of Problem~\ref{pr:resil_active_inf_acq} with the aforementioned constraints, and the monotone objective function~\cite{jawaid2015submodularity}:
\begin{equation*}
\metric \triangleq \frac{1}{T}\sum_{t=1}^{T} \log \det (\Sigma_t), \nonumber
\end{equation*}
where $\Sigma_{t+1} = \rho_{t+1}^{e}(\rho_t^p(\Sigma_t),x_{t+1})$ is the Kalman filtering Riccati map~\cite{atanasov2014information}.\footnote{We remark that the problem scenario is dependent on a prior distribution of the target's initial conditions $y_{0}$ and $\Sigma_{0|0}$. Notwithstanding, if a prior distribution is unknown, an exploration strategy can be incorporated to find the targets by placing exploration landmarks at the map frontiers \cite{atanasov2015decentralized}.}  We use the subroutines described in \cite{schlotfeldt2018anytime} and \cite{atanasov2015decentralized} for the step 1 and step 4 of Algorithm~\ref{alg:dec_resil_coord_decent}, respectively.

\subsection{Simulations on multi-target tracking with mobile robots}\label{subsec:sim}

\begin{figure}[t]
   \subfloat[][]{ \includegraphics[width=0.49\columnwidth]{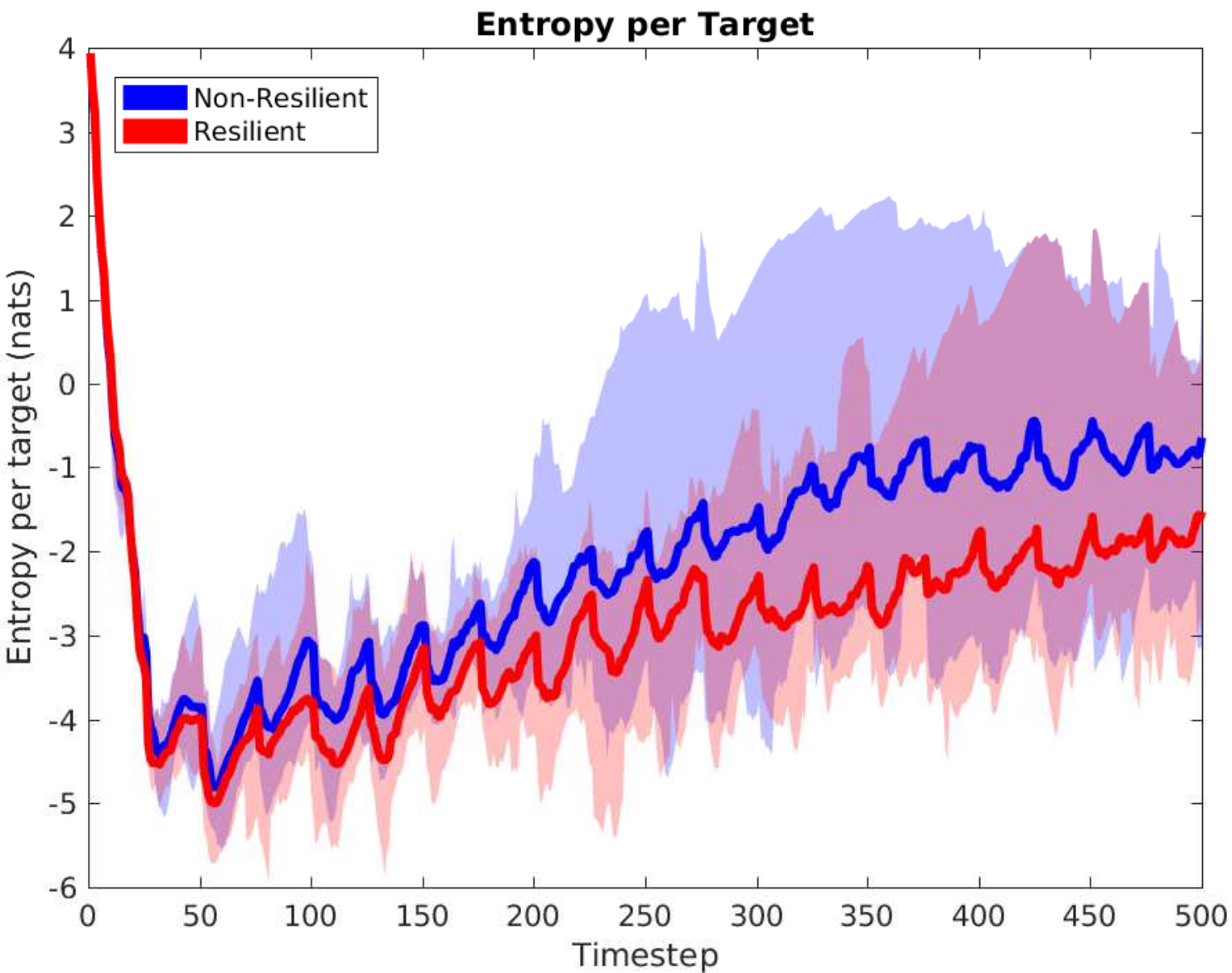}} 
   \subfloat[][]{ \includegraphics[width=0.49\columnwidth]{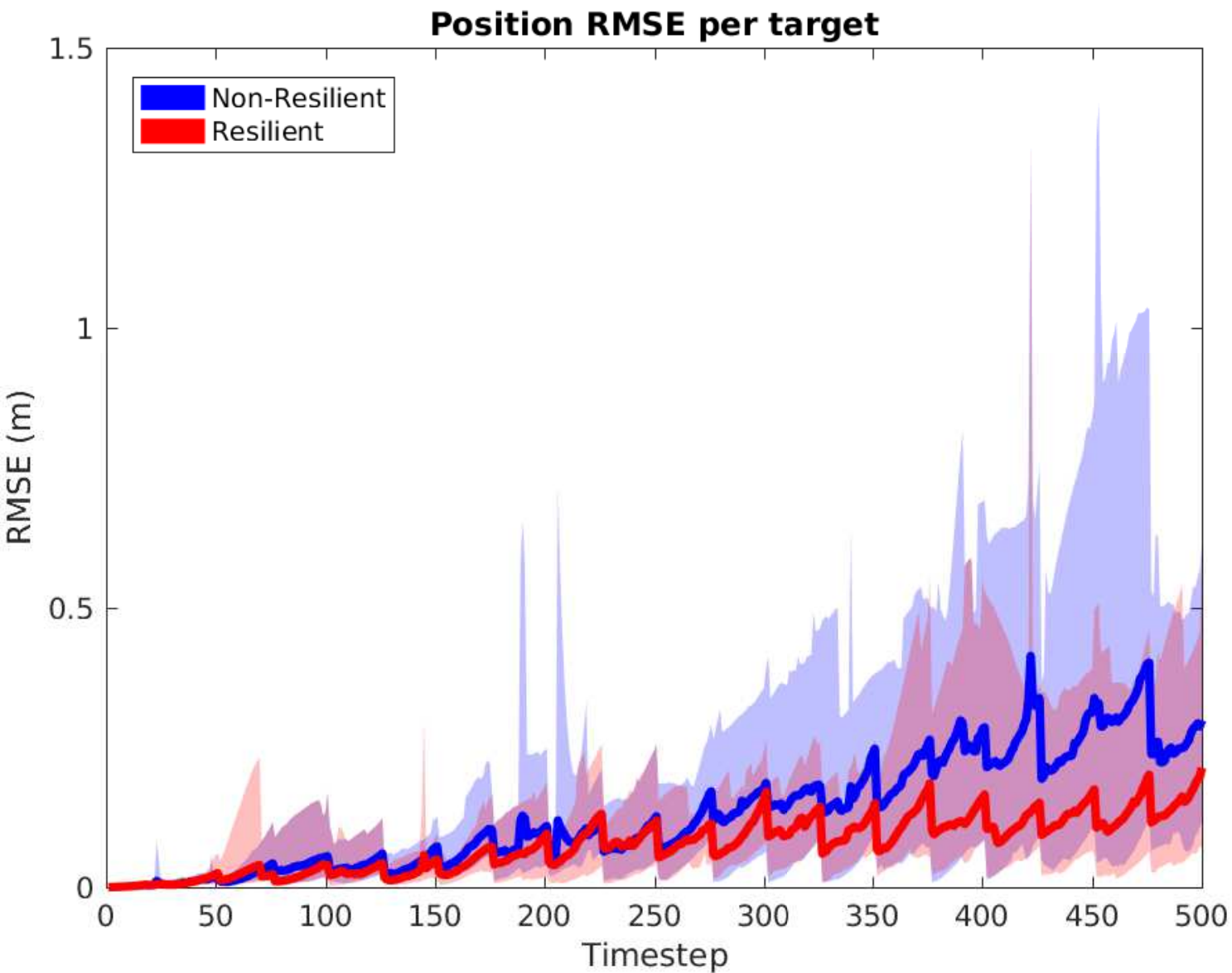}} \\
      \subfloat[][]{ \includegraphics[width=0.49\columnwidth]{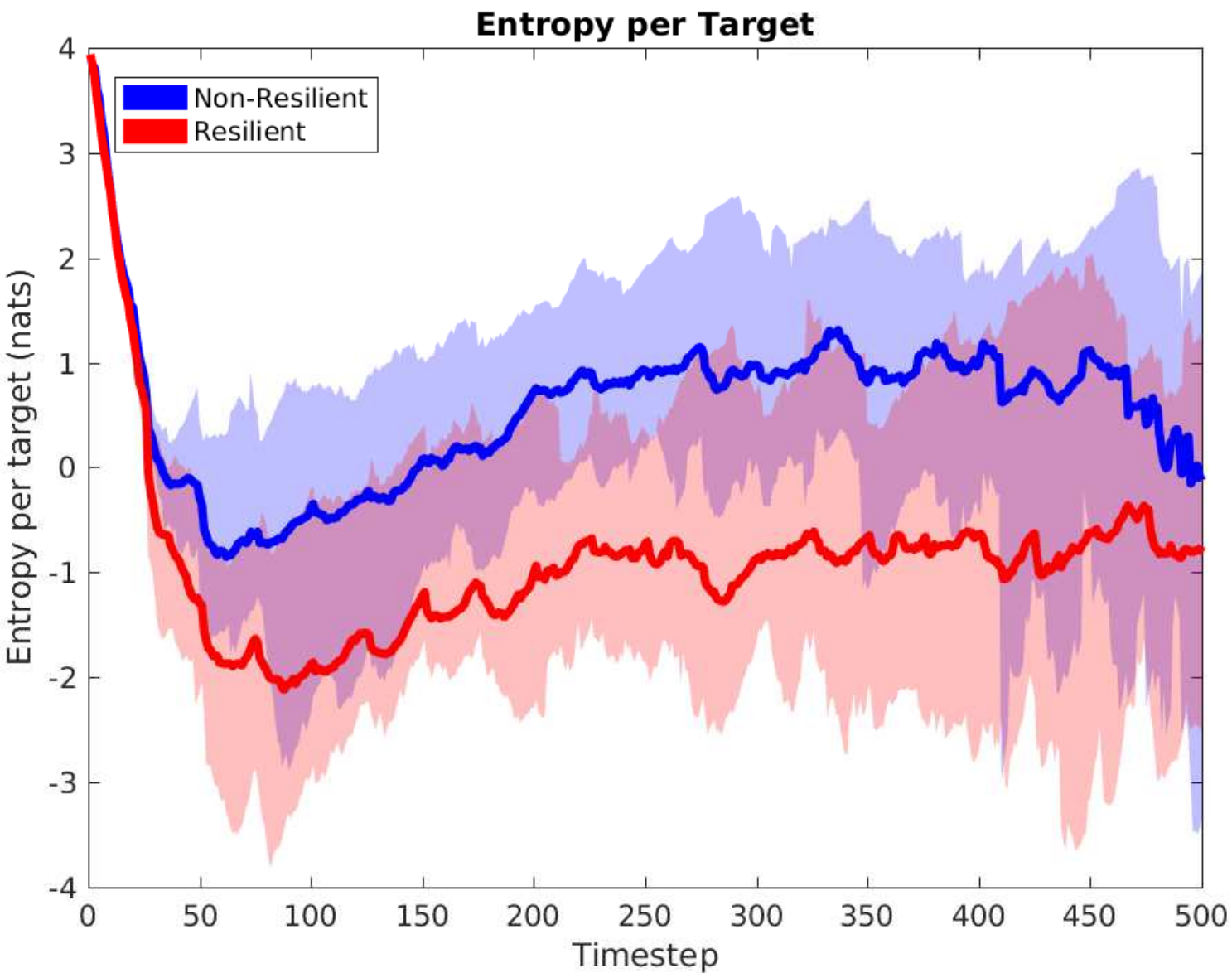}} 
         \subfloat[][]{ \includegraphics[width=0.49\columnwidth]{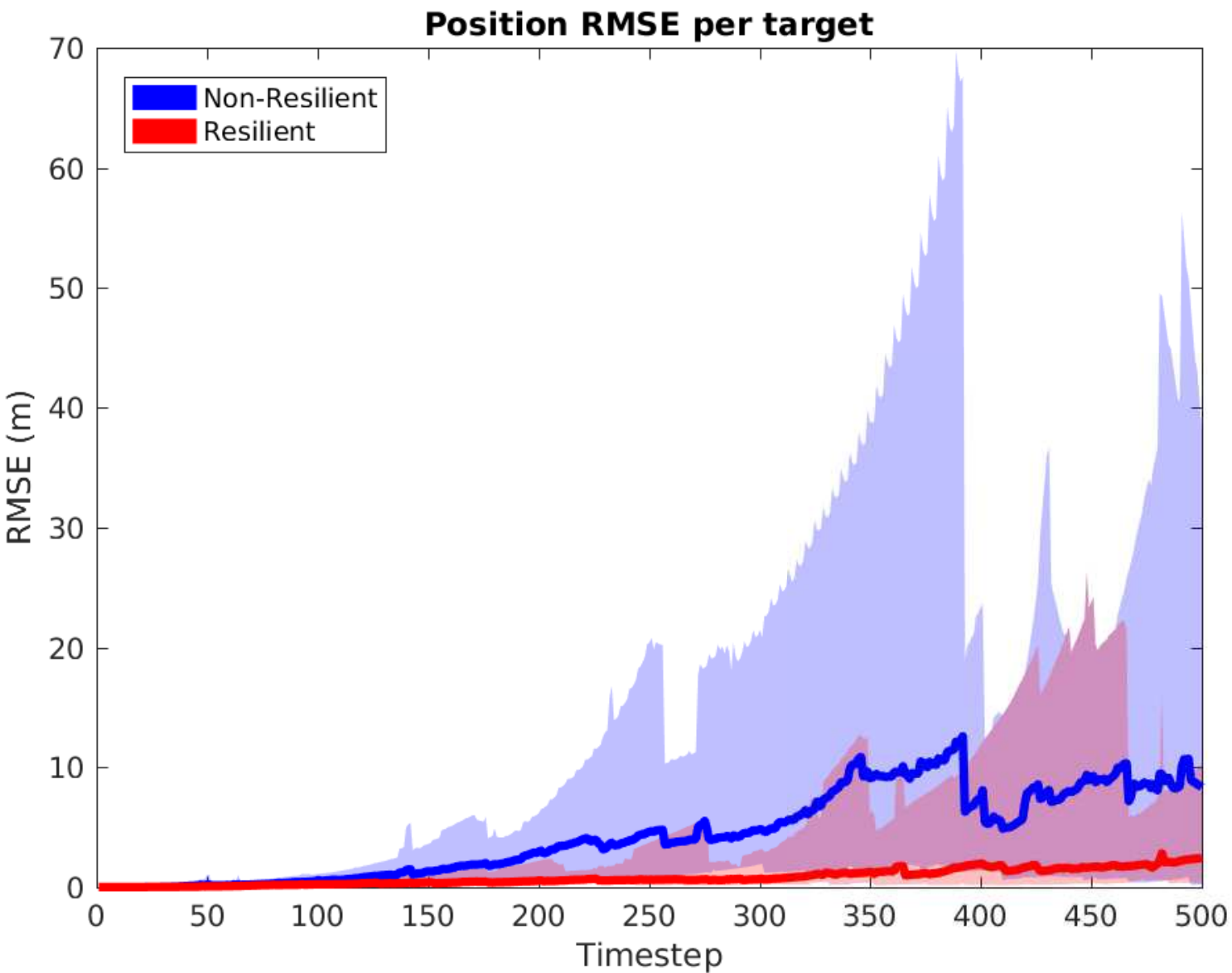}}
   
   \caption{\small The figures depict the average entropy and position RMSE (root mean square error) per target, averaged over the robots. Figs. (a-b) were obtained from a simulation with 10 robots, 10 targets, with 2 jamming attacks. Figs. (c-d) have the same configuration but up to 6 jamming attacks. The blue colors correspond to the non-resilient algorithm, and the red colors correspond to the resilient algorithm. The shaded regions are the spread between the minimum and maximum values of the information measure, and the solid lines are the mean value. {The plots are the aggregate of ten trials, each executed over 500 time-steps.}}
      \label{fig:sim_plots}
   \end{figure}

\indent We use simulations to evaluate the performance of our Algorithm~\ref{alg:dec_resil_coord_decent} across different scenarios. In particular, we vary the number of robots, $n$, the number of targets $M$, and the number of attacks~$\alpha$. In each of these scenarios we compare the performance of the resilient Algorithm~\ref{alg:dec_resil_coord_decent} with that of the non-resilient algorithm \textit{coordinate descent}~\cite[Section~IV]{atanasov2015decentralized}. To this end, we consider two information performance measures: the average entropy and average root mean square error (RMSE) per target, averaged over the robots in the team. \\ 
\indent We describe the parameters of the simulation: the robots and targets in the environment are restricted to move inside a 64x64 meter environment, as in Fig.~\ref{fig:sim_ex}. For the evaluation, we fix the initial positions of both the robots and targets, and the robots are given a prior distribution of the targets before starting the simulation. The targets start with a zero velocity, and in the event that a target leaves the environment its velocity is reflected to remain in bounds. Across all simulations we fix the remaining parameters as follows: $T=25$, $\tau= 0.5$, $r_{sense}=10$, $\psi=94^{\circ}$, $\sigma_r=.15$m, $\sigma_b = 5^{\circ}$, $q=.001$. Finally, we run Algorithm \ref{pr:resil_active_inf_acq} in a receding horizon fashion every $T$ time-steps, for a total of 500 steps, and average each configuration over 10 trials. The robots are forced to execute the entire $T$-step trajectory without re-planning, due to the jamming attack that occurs at the onset of every planning phase. Our results are depicted in Fig.~\ref{fig:sim_plots} and Table~\ref{fig:sim_table}. \\
\indent We observe in Fig. \ref{fig:sim_plots} that the performance of the resilient Algorithm~\ref{alg:dec_resil_coord_decent} is superior both with respect to the average entropy and the RMSE per target. Importantly, as the number of jamming attacks grows, the Algorithm~\ref{alg:dec_resil_coord_decent}'s superiority becomes more pronounced, and for the non-resilient algorithm the peaks in RMSE error grow much larger.

Table~\ref{fig:sim_table} suggests that the resilient Algorithm~\ref{alg:dec_resil_coord_decent} achieves a lower average error than the non-resilient algorithm, and, crucially, is highly effective in reducing the peak estimation error; in particular, Algorithm~\ref{alg:dec_resil_coord_decent} achieves a performance that is 2 to 30 times better in comparison to the performance achieved by the non-resilient algorithm. 
% Intuitively, this is because the Algorithm~\ref{alg:dec_resil_coord_decent} is conservative in guarding against a worst-case attack. 
We also observe that the impact of Algorithm~\ref{alg:dec_resil_coord_decent} is most prominent when the number of attacks is large relative to the size of the robot team.

\begin{table}[]
\centering
%\caption{Simulation Results}
\label{my-label}
\begin{tabular}{lllll}
                                                        & \multicolumn{2}{l}{Mean RMSE (m)}            & \multicolumn{2}{l}{Peak RMSE (m)}                                 \\ \hline
\multicolumn{1}{|l|}{}                                  & NR & \cellcolor[HTML]{C0C0C0}Resilient & NR & \multicolumn{1}{l|}{\cellcolor[HTML]{C0C0C0}Resilient} \\ \hline
\rowcolor[HTML]{EFEFEF} 
\multicolumn{1}{|l|}{\cellcolor[HTML]{EFEFEF}$n=5$, $M=10$} & \multicolumn{2}{l}{\cellcolor[HTML]{EFEFEF}} & \multicolumn{2}{l|}{\cellcolor[HTML]{EFEFEF}}                     \\
\multicolumn{1}{|l|}{$\alpha=1$}                        & 0.28     & \cellcolor[HTML]{C0C0C0}0.19      & 9.62     & \multicolumn{1}{l|}{\cellcolor[HTML]{C0C0C0}2.09}      \\
\multicolumn{1}{|l|}{$\alpha=2$}                        & 1.47     & \cellcolor[HTML]{C0C0C0}0.68      & 26.07    & \multicolumn{1}{l|}{\cellcolor[HTML]{C0C0C0}15.71}     \\
\multicolumn{1}{|l|}{$\alpha=4$}                        & 10.67    & \cellcolor[HTML]{C0C0C0}4.9       & 225.47   & \multicolumn{1}{l|}{\cellcolor[HTML]{C0C0C0}103.82}    \\
\rowcolor[HTML]{EFEFEF} 
\multicolumn{1}{|l|}{\cellcolor[HTML]{EFEFEF}$n=10$, $M=5$}  & \multicolumn{2}{l}{\cellcolor[HTML]{EFEFEF}} & \multicolumn{2}{l|}{\cellcolor[HTML]{EFEFEF}}                     \\
\multicolumn{1}{|l|}{$\alpha=2$}                        & 0.35     & \cellcolor[HTML]{C0C0C0}0.14      & 57.65    & \multicolumn{1}{l|}{\cellcolor[HTML]{C0C0C0}1.87}      \\
\multicolumn{1}{|l|}{$\alpha=4$}                        & 0.39     & \cellcolor[HTML]{C0C0C0}0.28      & 6.66     & \multicolumn{1}{l|}{\cellcolor[HTML]{C0C0C0}3.17}      \\
\multicolumn{1}{|l|}{$\alpha=6$}                        & 2.07     & \cellcolor[HTML]{C0C0C0}0.65      & 93.27    & \multicolumn{1}{l|}{\cellcolor[HTML]{C0C0C0}15.63}     \\
\rowcolor[HTML]{EFEFEF} 
\multicolumn{1}{|l|}{\cellcolor[HTML]{EFEFEF}$n=10$, $M=10$} & \multicolumn{2}{l}{\cellcolor[HTML]{EFEFEF}} & \multicolumn{2}{l|}{\cellcolor[HTML]{EFEFEF}}                     \\
\multicolumn{1}{|l|}{$\alpha=2$}                        & 0.13     & \cellcolor[HTML]{C0C0C0}0.08      & 1.4      & \multicolumn{1}{l|}{\cellcolor[HTML]{C0C0C0}1.32}      \\
\multicolumn{1}{|l|}{$\alpha=4$}                        & 0.24     & \cellcolor[HTML]{C0C0C0}0.23      & 4.19     & \multicolumn{1}{l|}{\cellcolor[HTML]{C0C0C0}2.66}      \\
\multicolumn{1}{|l|}{$\alpha=6$}                        & 4.39     & \cellcolor[HTML]{C0C0C0}1.2       & 69.77    & \multicolumn{1}{l|}{\cellcolor[HTML]{C0C0C0}26.4}      \\ \hline
\end{tabular}
\caption{\small The table depicts the estimation performance, measured by average and peak RMSE per tracked target, for a variety of configurations. The number $n$ denotes the number of mobile sensors, (i.e., $n=|\calV|$), $M$ denotes the number of moving targets, and $\alpha$ denotes the number of failures. NR denotes the non-resilient algorithm, while Resilient is Algorithm~\ref{alg:dec_resil_coord_decent}. All results are across 500 timesteps, averaged over ten trials per configuration.}
 \label{fig:sim_table}
\end{table}

\subsection{Experiments on multi-target tracking with mobile robots}\label{subsec:exp}
\begin{figure}[t]
\centering
\includegraphics[width=0.8\linewidth]{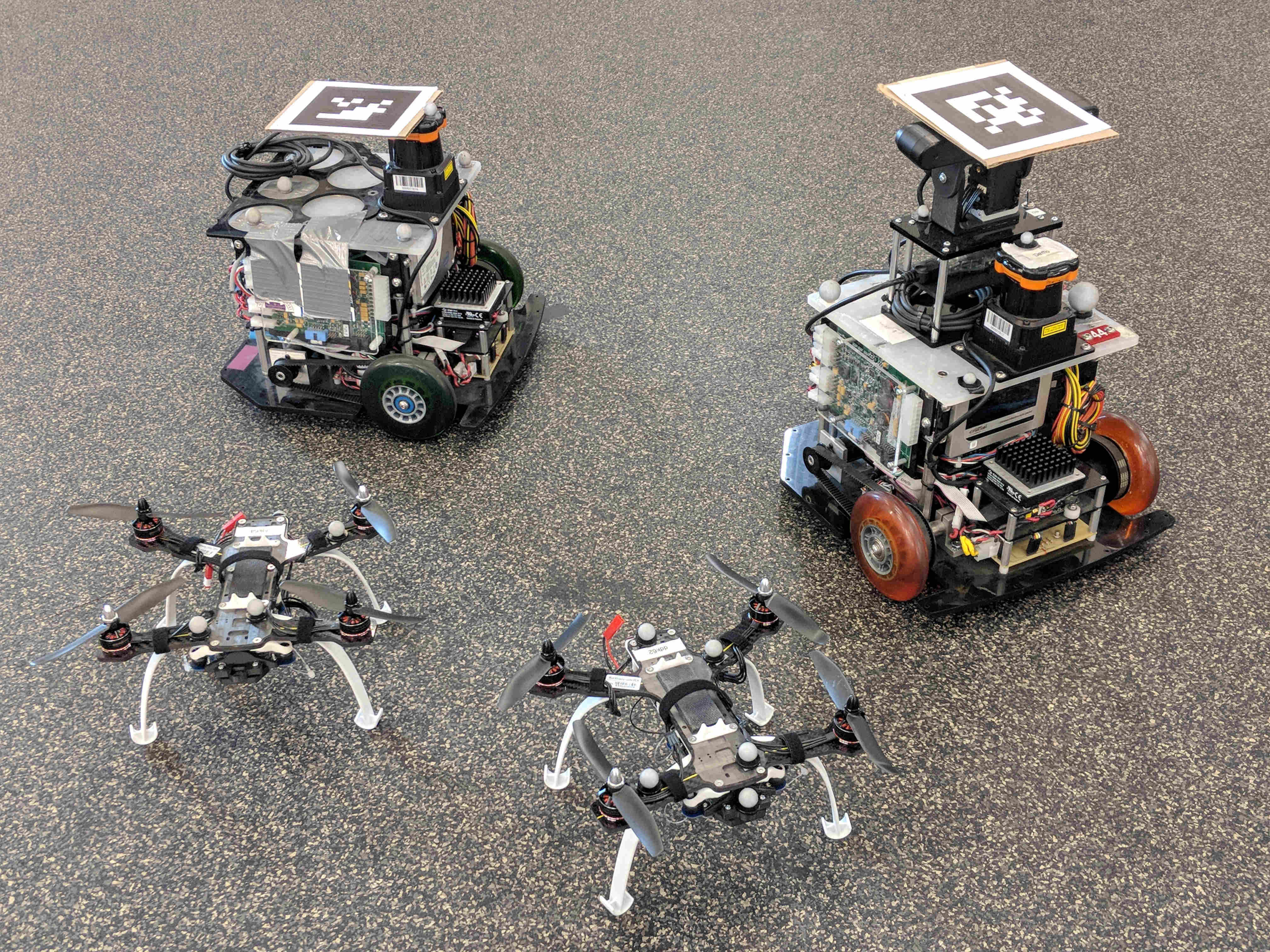}
\caption{\small The experimental setup with two quad-rotors equipped with Qualcomm Flight\texttrademark\!, and two Scarabs as ground targets.}
\label{fig:hw_ex}
\end{figure}

We implement Algorithm~\ref{alg:dec_resil_coord_decent} in a multi-UAV scenario with two quadrotors tracking the positions of two static ground targets, shown in Fig.~\ref{fig:hw_ex}.  The UAV trajectories are computed off-board but in \textit{real-time} on a laptop with an Intel Core i7 CPU. The UAVs are localized using the Vicon Motion Capture system. The UAVs are quad-rotors equipped with Qualcomm Flight\texttrademark\!. The UAVs use Vicon pose estimates to generate noisy measurements corresponding to a downward facing camera which has a $360^{\circ}$ field-of-view, and a $1$ meter sensing radius. The UAVs move in a 4x8 meter testing laboratory environment with no obstacles. One robot is jammed at all times.

The goal of the hardware experiments is to acquire a visual interpretation of the properties of the trajectories designed using the resilient Algorithm~\ref{alg:dec_resil_coord_decent}. To isolate the effect of resilience, we simplify the problem to static targets (i.e. stationary) and to the smallest possible team, i.e., 2 robots.  

\begin{figure}[t]
   \centering
   \subfloat[]{ \includegraphics[width=.85\columnwidth]{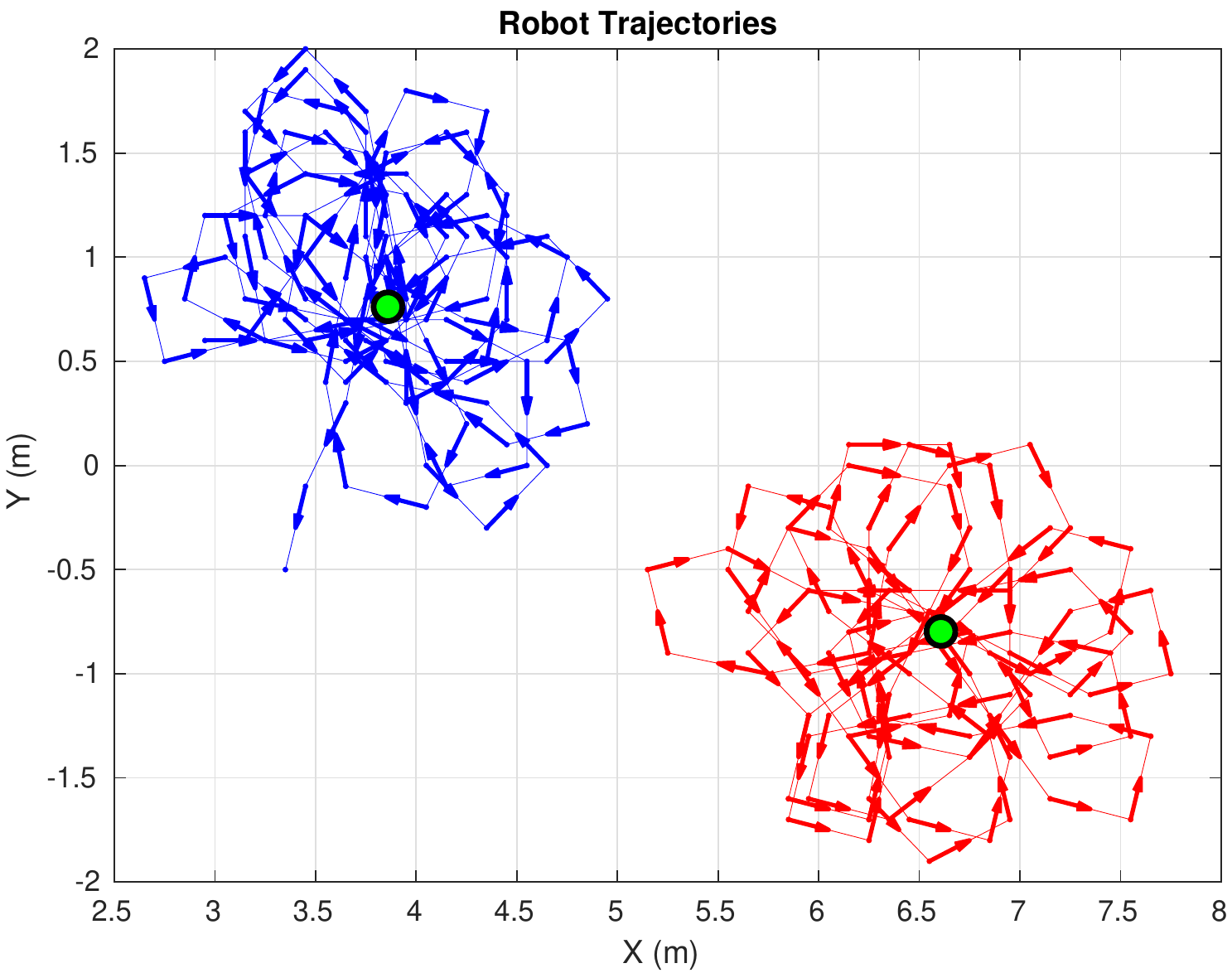}} \\
   \vspace*{-2ex}
   \subfloat[]{ \includegraphics[width=.85\columnwidth]{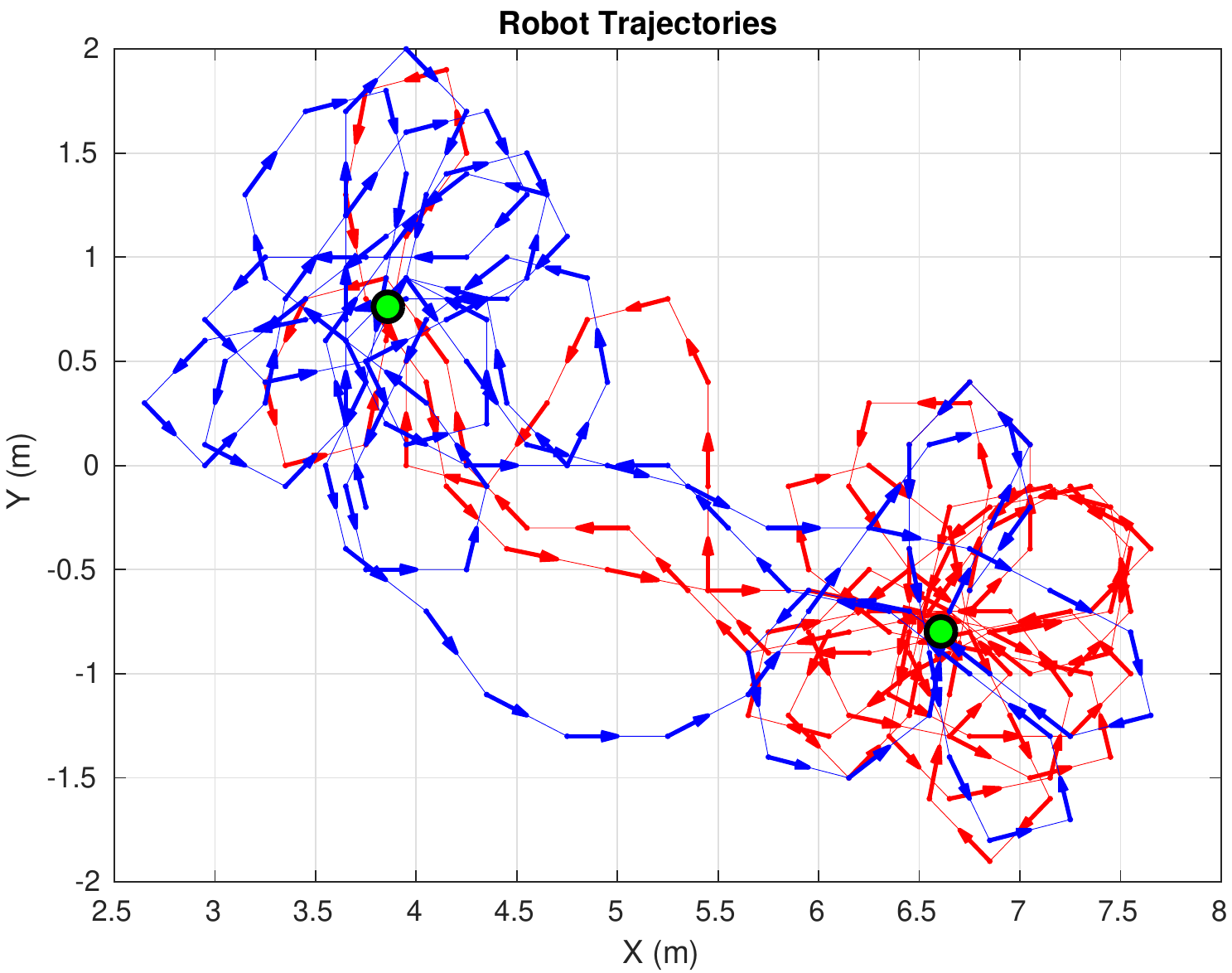}}
   \caption{\small The plot in (a) depicts the experimental robot trajectories in the non-resilient algorithm. The figure in (b) depicts the resilient algorithm. The targets are in green.}
      \label{fig:exp_plots}
   \end{figure}

We observe from the experiments that the trajectories planned by the UAVs under the non-resilient algorithm stick to the target they are closest to, whereas under the resilient Algorithm~\ref{alg:dec_resil_coord_decent}, the UAVs switch amongst the two targets (Fig. \ref{fig:exp_plots}). Intuitively, the reason is that the resilient algorithm always assumes that one of the robots will fail, in which case the optimal strategy for one UAV is to track two targets is to switch amongst the targets, whereas the non-resilient algorithm assumes that none of the robots will fail, in which case the optimal strategy for two UAVs is to allocate themselves to the closest target. 
When there is the possibility of one UAV failing, switching amongst the targets is preferable, since both robots have information about both targets.

%\input{tex/Conclusion.tex}
% % % % % % % % % % % % % % % % % % % % % % % % % % % % % % % % % % % % % % % % % % % % % % % % % % % % % % %
% % Section % % % %
\section{Concluding remarks \& Future work} \label{sec:con}
% % % % % % % % % % % % % % % % % % % % % % % % % % % % % % % % % % % % % % % % % % % % % 

We made the first steps to ensure the success of critical active information gathering tasks against failures and denial-of-service attacks, per Problem~\ref{pr:resil_active_inf_acq}.  In particular,  we provided the first algorithm for Problem~\ref{pr:resil_active_inf_acq}, and proved it guarantees 
 {near-optimal performance against system-wide failures}, even with minimal robot communication.
We~motivated the need for resilient active information gathering, and showcased the success of our algorithm, with simulated and real-world experiments in a series of {multi-robot target tracking} scenarios.

This paper opens a number of avenues for future research, both in theory and in applications.
Future work in theory includes the resilient design of the robot's communication network against network-wide failures, to balance the trade-off between~\textit{minimal communication}
% (which is necessitated in resource constrained environments, e.g., of limited bandwidth or of limited battery) 
and~\textit{connectedness}, which is necessitated in scenarios that are both resource constrained (e.g., where bandwidth or battery is limited), and failure-prone (e.g., where attacks can disrupt communication links). Future work in applications includes the experimental testing of resilient active information gathering with mobile robots in environmental monitoring, search and rescue scenarios, and simultaneous localization and mapping.% (SLAM).

\section{Acknowledgements}
We thank Nikolay A.~Atanasov for inspiring discussions. 

\bibliographystyle{IEEEtran}
\bibliography{references}

\begin{thebibliography}{10}
\providecommand{\url}[1]{#1}
\csname url@samestyle\endcsname
\providecommand{\newblock}{\relax}
\providecommand{\bibinfo}[2]{#2}
\providecommand{\BIBentrySTDinterwordspacing}{\spaceskip=0pt\relax}
\providecommand{\BIBentryALTinterwordstretchfactor}{4}
\providecommand{\BIBentryALTinterwordspacing}{\spaceskip=\fontdimen2\font plus
\BIBentryALTinterwordstretchfactor\fontdimen3\font minus
  \fontdimen4\font\relax}
\providecommand{\BIBforeignlanguage}[2]{{%
\expandafter\ifx\csname l@#1\endcsname\relax
\typeout{** WARNING: IEEEtran.bst: No hyphenation pattern has been}%
\typeout{** loaded for the language `#1'. Using the pattern for}%
\typeout{** the default language instead.}%
\else
\language=\csname l@#1\endcsname
\fi
#2}}
\providecommand{\BIBdecl}{\relax}
\BIBdecl

\bibitem{michini2014robotic}
M.~Michini, M.~A. Hsieh, E.~Forgoston, and I.~B. Schwartz, ``Robotic tracking
  of coherent structures in flows,'' \emph{IEEE Transactions on Robotics},
  vol.~30, no.~3, pp. 593--603, 2014.

\bibitem{nieto2013multi}
C.~Nieto-Granda, J.~G. {Rogers III}, and H.~Christensen, ``Multi-robot
  exploration strategies for tactical tasks in urban environments,'' in
  \emph{SPIE Defense, Security, and Sensing}.\hskip 1em plus 0.5em minus
  0.4em\relax International Society for Optics and Photonics, 2013, pp.
  87\,410B--87\,410B.

\bibitem{kumar2012opportunities}
V.~Kumar and N.~Michael, ``Opportunities and challenges with autonomous micro
  aerial vehicles,'' \emph{The International Journal of Robotics Research},
  vol.~31, no.~11, pp. 1279--1291, 2012.

\bibitem{karaman2012high}
S.~Karaman and E.~Frazzoli, ``High-speed flight in an ergodic forest,'' in
  \emph{IEEE Intern.~Confer.~on Robotics and Automation}, 2012, pp. 2899--2906.

\bibitem{cadena2016past}
C.~Cadena, L.~Carlone, H.~Carrillo, Y.~Latif, D.~Scaramuzza, J.~Neira, I.~Reid,
  and J.~J. Leonard, ``Past, present, and future of simultaneous localization
  and mapping: Toward the robust-perception age,'' \emph{IEEE Transactions on
  Robotics}, vol.~32, no.~6, pp. 1309--1332, 2016.

\bibitem{8206030}
T.~Cieslewski, E.~Kaufmann, and D.~Scaramuzza, ``Rapid exploration with
  multi-rotors: A frontier selection method for high speed flight,'' in
  \emph{IEEE/RSJ Int.~Conf.~on Intel.~Robots and Systems}, 2017, pp.
  2135--2142.

\bibitem{8255576}
M.~Santos, Y.~Diaz-Mercado, and M.~Egerstedt, ``Coverage control for multirobot
  teams with heterogeneous sensing capabilities,'' \emph{IEEE Robotics and
  Automation Letters}, vol.~3, no.~2, pp. 919--925, 2018.

\bibitem{krause2008optimizing}
A.~Krause, ``Optimizing sensing: Theory and applications,'' Ph.D. dissertation,
  Carnegie Mellon University, 2008.

\bibitem{williams2007information}
J.~L. Williams, ``Information theoretic sensor management,'' Ph.D.
  dissertation, Massachusetts Institute of Technology, 2007.

\bibitem{tzoumas2016near}
V.~Tzoumas, A.~Jadbabaie, and G.~J. Pappas, ``Near-optimal sensor scheduling
  for batch state estimation,'' in \emph{IEEE 55th Conference on Decision and
  Control}, 2016, pp. 2695--2702.

\bibitem{hoffmann2010mobile}
G.~M. Hoffmann and C.~J. Tomlin, ``Mobile sensor network control using mutual
  information methods and particle filters,'' \emph{IEEE Transactions on
  Automatic Control}, vol.~55, no.~1, pp. 32--47, 2010.

\bibitem{julian2012distributed}
B.~J. Julian, M.~Angermann, M.~Schwager, and D.~Rus, ``Distributed robotic
  sensor networks: An information-theoretic approach,'' \emph{The
  Inter.~Journal of Robotics Research}, vol.~31, no.~10, pp. 1134--1154, 2012.

\bibitem{dames2012decentralized}
P.~Dames, M.~Schwager, V.~Kumar, and D.~Rus, ``A decentralized control policy
  for adaptive information gathering in hazardous environments,'' in \emph{IEEE
  Conference on Decision and Control}, 2012, pp. 2807--2813.

\bibitem{dames2015autonomous}
P.~Dames and V.~Kumar, ``Autonomous localization of an unknown number of
  targets using teams of mobile sensors,'' \emph{IEEE Trans.~on Autom.~Science
  and Eng.}, vol.~12, no.~3, pp. 850--864, 2015.

\bibitem{charrow2014approximate}
B.~Charrow, V.~Kumar, and N.~Michael, ``Approximate representations for
  multi-robot control policies that maximize mutual information,''
  \emph{Autonomous Robots}, vol.~37, no.~4, pp. 383--400, 2014.

\bibitem{chung2006decentralized}
T.~H. Chung, J.~W. Burdick, and R.~M. Murray, ``A decentralized motion
  coordination strategy for dynamic target tracking,'' in \emph{IEEE
  International Conference on Robotics and Automation}, 2006, pp. 2416--2422.

\bibitem{kreucher2005information}
C.~M. Kreucher, ``An information-based approach to sensor resource
  allocation,'' Ph.D. dissertation, University of Michigan, 2005.

\bibitem{atanasov2014information}
N.~Atanasov, J.~Le~Ny, K.~Daniilidis, and G.~J. Pappas, ``Information
  acquisition with sensing robots: {A}lgorithms and error bounds,'' in
  \emph{IEEE International Confer.~on Robotics and Automation}, 2014, pp.
  6447--6454.

\bibitem{schlotfeldt2018anytime}
B.~Schlotfeldt, D.~Thakur, N.~Atanasov, V.~Kumar, and G.~J. Pappas, ``Anytime
  planning for decentralized multi-robot active information gathering,''
  \emph{IEEE Robotics and Automation Letters}, 2018.

\bibitem{atanasov2015decentralized}
N.~Atanasov, J.~Le~Ny, K.~Daniilidis, and G.~J. Pappas, ``Decentralized active
  information acquisition: Theory and application to multi-robot slam,'' in
  \emph{IEEE Int.~Conf.~on Rob.~and Autom.}, 2015, pp. 4775--4782.

\bibitem{orlin2016robust}
J.~B. Orlin, A.~S. Schulz, and R.~Udwani, ``Robust monotone submodular function
  maximization,'' in \emph{Inter. Conference on Integer Programming and
  Combinatorial Optimization}, 2016, pp. 312--324.

\bibitem{tzoumas2017resilient}
V.~Tzoumas, K.~Gatsis, A.~Jadbabaie, and G.~J. Pappas, ``Resilient monotone
  submodular function maximization,'' in \emph{IEEE Conference on Decision and
  Control}, 2017, pp. 1362--1367.

\bibitem{tzoumas2018resilientSequential}
V.~{Tzoumas}, A.~{Jadbabaie}, and G.~J. {Pappas}, ``{Resilient Monotone
  Sequential Maximization},'' \emph{ArXiv e-prints: 1803.07954}.

\bibitem{bertsekas2005dynamic}
D.~P. Bertsekas, \emph{Dynamic programming and optimal control, Vol. I}.\hskip
  1em plus 0.5em minus 0.4em\relax Athena Scientific, 2005.

\bibitem{Feige:1998:TLN:285055.285059}
U.~Feige, ``A threshold of $ln(n)$ for approximating set cover,'' \emph{Journal
  of the ACM}, vol.~45, no.~4, pp. 634--652, 1998.

\bibitem{nemhauser78analysis}
G.~Nemhauser, L.~Wolsey, and M.~Fisher, ``An analysis of approximations for
  maximizing submodular set functions -- {I},'' \emph{Mathematical
  Programming}, vol.~14, no.~1, pp. 265--294, 1978.

\bibitem{conforti1984curvature}
M.~Conforti and G.~Cornu\'{e}jols, ``Submodular set functions, matroids and the
  greedy algorithm,'' \emph{Discrete Applied Mathematics}, vol.~7, no.~3, pp.
  251 -- 274, 1984.

\bibitem{sviridenko2017optimal}
M.~Sviridenko, J.~Vondr{\'a}k, and J.~Ward, ``Optimal approximation for
  submodular and supermodular optimization with bounded curvature,''
  \emph{Math.~of Operations Research}, vol.~42, no.~4, pp. 1197--1218, 2017.

\bibitem{jawaid2015submodularity}
S.~T. Jawaid and S.~L. Smith, ``Submodularity and greedy algorithms in sensor
  scheduling for linear dynamical systems,'' \emph{Automatica}, vol.~61, pp.
  282--288, 2015.

\end{thebibliography}

 \appendices

 \section{Preliminary lemmas and definitions}\label{app:prelim}

\myParagraph{Notation}
In the appendix we use the following notation
to support the proofs in this paper: in particular, consider a finite ground set $\mathcal{V}$ and a set function $f:2^\mathcal{V}\mapsto \mathbb{R}$. Then, for any set $\mathcal{X}\subseteq \mathcal{V}$ and any set $\mathcal{X}'\subseteq \mathcal{V}$, the symbol $f(\mathcal{X}|\mathcal{X}')$ denotes the marginal value  $f(\mathcal{X}\cup\mathcal{X}')-f(\mathcal{X}')$.  Moreover, the symbol $\kappa_f$ is the total curvature of $f$ (Definition~\ref{def:curvature}), and the symbol $c_f$ is the total curvature of $f$ (Definition~\ref{def:total_curvature}).

\medskip

This appendix contains lemmas that will support the proof of Theorem~\ref{th:per_alg_dec_resil_coord_decent} in this paper; moreover, it contains a generalized description of the algorithm \textit{coordinate descent}~\cite[Section~IV]{atanasov2015decentralized} (to any non-decreasing information objective function in the active robot set), and a lemma, which will support the proof of Proposition~\ref{prop:alg_per_with_coordinate_descent} in this paper.

\subsection{Lemmas that support the proof of Theorem~\ref{th:per_alg_dec_resil_coord_decent}}

The proof of the lemmas is also found in~\cite{tzoumas2017resilient,tzoumas2018resilientSequential}.

\begin{mylemma}\label{lem:non_total_curvature}
Consider a finite ground set $\mathcal{V}$ and a non-decreasing and submodular set function $f:2^\mathcal{V}\mapsto \mathbb{R}$ such that $f$ is non-negative and $f(\emptyset)=0$. For any $\mathcal{A}\subseteq \mathcal{V}$, it~is:
\begin{equation*}
f(\mathcal{A})\geq (1-\kappa_f)\sum_{a \in \mathcal{A}}f(a).
\end{equation*}
\end{mylemma}
\paragraph*{Proof of Lemma~\ref{lem:non_total_curvature}} Let $\mathcal{A}=\{a_1,a_2,\ldots, a_{|{\cal A}|}\}$. We prove Lemma~\ref{lem:curvature} by proving the following two inequalities: 
\begin{align}
f(\mathcal{A})&\geq \sum_{i=1}^{|{\cal A}|} f(a_i|\mathcal{V}\setminus\{a_i\}),\label{ineq:aux_5}\\
\sum_{i=1}^{|{\cal A}|} f(a_i|\mathcal{V}\setminus\{a_i\})&\geq (1-\kappa_f)\sum_{i=1}^{|{\cal A}|} f(a_i)\label{ineq:aux_6}. 
\end{align} 

We begin with the proof of ineq.~\eqref{ineq:aux_5}: 
\begin{align}
f(\mathcal{A})&=f(\mathcal{A}|\emptyset)\label{ineq:aux_9}\\
&\geq f(\mathcal{A}|\mathcal{V}\setminus \mathcal{A})\label{ineq:aux_10}\\
&= \sum_{i=1}^{|{\cal A}|}f(a_i|\mathcal{V}\setminus\{a_i,a_{i+1},\ldots,a_{|{\cal A}|}\})\label{ineq:aux_11}\\
&\geq \sum_{i=1}^{|{\cal A}|}f(a_i|\mathcal{V}\setminus\{a_i\}),\label{ineq:aux_12}
\end{align}
where ineqs.~\eqref{ineq:aux_10} to~\eqref{ineq:aux_12} hold for the following reasons: ineq.~\eqref{ineq:aux_10} is implied by eq.~\eqref{ineq:aux_9} because $f$ is submodular and $\emptyset\subseteq \mathcal{V}\setminus \mathcal{A}$; eq.~\eqref{ineq:aux_11} holds since for any sets $\mathcal{X}\subseteq \mathcal{V}$ and $\mathcal{Y}\subseteq \mathcal{V}$ it is $f(\mathcal{X}|\mathcal{Y})=f(\mathcal{X}\cup \mathcal{Y})-f(\mathcal{Y})$, and it also  $\{a_1,a_2,\ldots, a_{|{\cal A}|}\}$ denotes the set $\mathcal{A}$; and ineq.~\eqref{ineq:aux_12} holds since $f$ is submodular and $\mathcal{V}\setminus\{a_i,a_{i+1},\ldots,a_{\mu}\} \subseteq \mathcal{V}\setminus\{a_i\}$.  These observations complete the proof of ineq.~\eqref{ineq:aux_5}.

We now prove ineq.~\eqref{ineq:aux_6} using the Definition~\ref{def:curvature} of $\kappa_f$, as follows: since $\kappa_f=1-\min_{v\in \mathcal{V}}\frac{f(v|\mathcal{V}\setminus\{v\})}{f(v)}$, it is implied that for all elements $v\in \mathcal{V}$ it is $ f(v|\mathcal{V}\setminus\{v\})\geq (1-\kappa_f)f(v)$.  Therefore, adding the latter inequality across all elements $a \in \calA$ completes the proof of ineq.~\eqref{ineq:aux_6}.
\hfill $\blacksquare$

\begin{mylemma}\label{lem:D3}
Consider any finite ground set $\mathcal{V}$, a non-decreasing and submodular function $f:2^\mathcal{V}\mapsto \mathbb{R}$ and non-empty sets $\calY, \calP \subseteq \calV$ such that for all elements $y \in \calY$ and all elements $p \in \calP$ it is $f(y)\geq f(p)$.  Then, it is:
\belowdisplayskip=-12pt\begin{equation*}
f(\calP|\calY)\leq |\calP|f(\calY).
\end{equation*}
\end{mylemma}
\paragraph*{Proof of Lemma~\ref{lem:D3}} Consider any element $y \in \calY$ (such an element exists since Lemma~\ref{lem:D3} considers that $\calY$ is non-empty); then, 
\begin{align}
f(\calP|\calY)&= f(\calP\cup\calY)-f(\calY)\label{aux1:1}\\
&\leq f(\calP)+f(\calY)-f(\calY)\label{aux1:2}\\
&= f(\calP)\nonumber\\
&\leq \sum_{p\in\calP}f(p)\label{aux1:4}\\
&\leq |\calP| \max_{p\in\calP} f(p)\nonumber\\
&\leq |\calP|  f(y)\label{aux1:7}\\
&\leq |\calP| f(\calY),\label{aux1:5}
\end{align}
where eq.~\eqref{aux1:1} to ineq.~\eqref{aux1:5} hold for the following reasons: eq.~\eqref{aux1:1} holds since for any sets $\mathcal{X}\subseteq \mathcal{V}$ and $\mathcal{Y}\subseteq \mathcal{V}$, it is $f(\mathcal{X}|\mathcal{Y})=f(\mathcal{X}\cup \mathcal{Y})-f(\mathcal{Y})$; ineq.~\eqref{aux1:2} holds since $f$ is submodular and, as a result, the submodularity Definition~\ref{def:sub} implies that for any set $\calA\subseteq\calV$ and $\calA'\subseteq\calV$, it is $f(\calA\cup \calA')\leq f(\calA)+f(\calA')$; ineq.~\eqref{aux1:4} holds for the same reason as ineq.~\eqref{aux1:2}; ineq.~\eqref{aux1:7} holds since or all elements $y \in \calY$ and all elements $p \in \calP$ it is $f(y)\geq f(p)$; finally, ineq.~\eqref{aux1:5} holds because $f$ is monotone and $y\in\calY$.
\hfill $\blacksquare$

\begin{mylemma}\label{lem:curvature}
Consider a finite ground set $\mathcal{V}$ and a non-decreasing set function $\function:2^\mathcal{V}\mapsto \mathbb{R}$ such that $\function$ is non-negative and $\function(\emptyset)=0$. For any set $\mathcal{A}\subseteq \mathcal{V}$ and any set $\mathcal{B}\subseteq \mathcal{V}$ such that $\calA \cap \calB=\emptyset$, it is:
\begin{equation*}
\function(\mathcal{A}\cup \mathcal{B})\geq (1-c_\function)\left(\function(\mathcal{A})+\sum_{b \in \mathcal{B}}\function(b)\right).
\end{equation*}
\end{mylemma}
\paragraph*{Proof of Lemma~\ref{lem:curvature}}
Let $\mathcal{B}=\{b_1, b_2, \ldots, b_{|\mathcal{B}|}\}$. Then, 
\begin{equation}
f(\mathcal{A}\cup \mathcal{B})=\function(\mathcal{A})+\sum_{i=1}^{|\mathcal{B}|}\function(b_i|\mathcal{A}\cup \{b_1, b_2, \ldots, b_{i-1}\}). \label{eq1:lemma_curvature}
\end{equation} 
In addition, Definition~\ref{def:total_curvature} of total curvature implies:
\begin{align}
\function(b_i|\mathcal{A}\cup \{b_1, b_2, \ldots, b_{i-1}\})&\geq (1-c_f)\function(b_i|\emptyset)\nonumber\\
&=(1-c_f)\function(b_i), \label{eq2:lemma_curvature}
\end{align} 
where the latter equation holds since $\function(\emptyset)=0$.
The proof is completed by substituting~\eqref{eq2:lemma_curvature} in~\eqref{eq1:lemma_curvature} and then taking into account that $\function(\mathcal{A})\geq (1-c_f)\function(\mathcal{A})$ since $0\leq c_\function\leq 1$. \hfill $\blacksquare$

\begin{mylemma}\label{lem:subratio}
Consider a finite ground set $\mathcal{V}$ and a non-decreasing set function $f:2^\mathcal{V}\mapsto \mathbb{R}$ such that $f$ is non-negative and $f(\emptyset)=0$. For any set $\mathcal{A}\subseteq \mathcal{V}$ and any set  $\mathcal{B}\subseteq \mathcal{V}$ such that $\mathcal{A}\setminus\mathcal{B}\neq \emptyset$, it is:
\begin{equation*}
f(\mathcal{A})+(1-c_f) f(\mathcal{B})\geq (1-c_f) f(\mathcal{A}\cup \mathcal{B})+f(\mathcal{A}\cap \mathcal{B}).
\end{equation*}
\end{mylemma}
\paragraph*{Proof of Lemma~\ref{lem:subratio}}
Let $\mathcal{A}\setminus\mathcal{B}=\{i_1,i_2,\ldots, i_r\}$, where $r=|\mathcal{A}-\mathcal{B}|$. From Definition~\ref{def:total_curvature} of total curvature $c_f$, for any $i=1,2, \ldots, r$, it is  $f(i_j|\mathcal{A} \cap \mathcal{B} \cup \{i_1, i_2, \ldots, i_{j-1}\})\geq (1-c_f) f(i_j|\mathcal{B} \cup \{i_1, i_2, \ldots, i_{j-1}\})$. Summing these $r$ inequalities,
$$f(\mathcal{A})-f(\mathcal{A}\cap \mathcal{B})\geq (1-c_f) \left(f(\mathcal{A}\cup \mathcal{B})-f(\mathcal{B})\right),$$
which implies the lemma. \hfill $\blacksquare$

\begin{mycorollary}\label{cor:ineq_from_lemmata}
Consider a finite ground set $\mathcal{V}$ and a non-decreasing set function $f:2^\mathcal{V}\mapsto \mathbb{R}$ such that $f$ is non-negative and $f(\emptyset)=0$. For any set $\mathcal{A}\subseteq \mathcal{V}$ and any set $\mathcal{B}\subseteq \mathcal{V}$ such that $\mathcal{A}\cap\mathcal{B}=\emptyset$, it is:
\begin{equation*}
f(\mathcal{A})+\sum_{b \in \mathcal{B}}f(b) \geq (1-c_f)  f(\mathcal{A}\cup \mathcal{B}).
\end{equation*}
\end{mycorollary}
\paragraph*{Proof of Corollary~\ref{cor:ineq_from_lemmata}}
 Let $\mathcal{B}=\{b_1,b_2,\ldots,b_{|\mathcal{B}|}\}$. 
\begin{align}
f(\mathcal{A})+\sum_{i=1}^{|\mathcal{B}|}f(b_i) &\geq (1-c_f) f(\mathcal{A})+\sum_{i=1}^{|\mathcal{B}|}f(b_i))\label{ineq:cor_aux1} \\
& \geq (1-c_f) f(\mathcal{A}\cup \{b_1\})+\sum_{i=2}^{|\mathcal{B}|}f(b_i)\nonumber\\
& \geq (1-c_f) f(\mathcal{A}\cup \{b_1,b_2\})+\sum_{i=3}^{|\mathcal{B}|}f(b_i)\nonumber\\
& \;\;\vdots \nonumber\\
& \geq (1-c_f) f(\mathcal{A}\cup \mathcal{B}),\nonumber
\end{align}
where~\eqref{ineq:cor_aux1} holds since $0\leq c_f\leq 1$, and the rest due to Lemma~\ref{lem:subratio} since $\mathcal{A}\cap\mathcal{B}=\emptyset$ implies $\mathcal{A}\setminus \{b_1\}\neq \emptyset$, $\mathcal{A}\cup \{b_1\}\setminus \{b_2\}\neq \emptyset$, $\ldots$, $\mathcal{A}\cup \{b_1,b_2,\ldots, b_{|\mathcal{B}|-1}\}\setminus \{b_{|\mathcal{B}|}\}\neq \emptyset$. 

\hfill $\blacksquare$

\subsection{Generalized Coordinate Descent and a lemma \\ that supports the proof of Proposition~\ref{prop:alg_per_with_coordinate_descent}}\label{app:description_coordinate}

In this section we generalize the proof in \cite{atanasov2015decentralized} that the algorithm \textit{coordinate descent} proposed therein guarantees for the information objective function of mutual information an approximation performance up to a multiplicative factor $1/2$ the optimal.  In particular, we extend the proof to \textit{any} non-decreasing and submodular information objective function, as well as to \textit{any} non-decreasing information objective function. 

The algorithm coordinate descent 
works as follows: consider an \textit{arbitrary} ordering of the robots in $\calV$, such that $\calV \equiv \{1,2,\ldots, n\}$, and suppose that robot $1$ chooses first its controls, without considering the other robots; in other words, robot $1$ solves the single robot version of Problem~\ref{pr:resil_active_inf_acq}, i.e. $\setP(\{1\}, 0)$, to obtain control inputs $u_{1:T}^{cd}(\{1\})$ such that:
\begin{equation} \label{eq:descent}
\begin{aligned}
u_{1:T}^{cd}(\{1\}) \in \mathop{\arg\min}_{\hat{u}_t \in \mathcal{U}_{1,t}, t=1,2,\ldots,T} J(\hat{u}_{1:T}).
\end{aligned}
\end{equation}
Afterwards, robot $1$ communicates its chosen control sequence to robot 2, and robot 2, given the control sequence of robot~1, computes its control input as follows:
\begin{equation}
\begin{aligned}
u_{1:T}^{cd}(\{2\}) \in \mathop{\arg\min}_{\hat{u}_t \in \mathcal{U}_{2,t}, t=1,2,\ldots,T} 
J(u_{1:T}^{cd}(\{1\}), \hat{u}_{1:T} ).
\end{aligned}
\end{equation}
This continues such that robot $i+1$ solves a single robot problem, given the control inputs from the robots $1,2,\ldots,i$:
\begin{equation}
\begin{aligned}
u_{1:T}^{cd}(\{i\}) \in \mathop{\arg\min}_{\hat{u}_t \in \mathcal{U}_{2,t}, t=1,2,\ldots,T} J(u_{1:T}^{cd}(\{1,2,\ldots,i\}), \hat{u}_{1:T} ).
\end{aligned}
\end{equation}

Notably, if we let $u^*_{1:T}(\{i\})$ be the control inputs for the $i$-th robot resulting from the optimal solution to the $n$ robot problem, then from the coordinate descent algorithm it is:
\begin{equation}\label{cd_policy}
\begin{aligned}
J(u^{cd}_{1:T}(\{1,2,\ldots,i\}), u^*_{1:T}(\{i\}) ) \leq J(u^{cd}_{1:T}(\{1,2,\ldots,i\}).
\end{aligned}
\end{equation}

\begin{mylemma}\label{lem:generalized_cd} \emph{\textbf{(Approximation performance of coordinate descent)}}
Consider a set of robots $\calV$, and an instance of problem $\setP(\calV, 0)$, per eq.~\eqref{def:resil_active_inf_acq}. Denote the optimal control inputs for problem $\setP(\calV, 0)$, across all robots and all times, by $u^*_{1:T}(\calV)$. The coordinate descent algorithm returns control inputs $u^{cd}_{1:T}(\calV)$, across all robots and all times, such that:
\begin{itemize}
    \item if the objective function $\metric$ is non-decreasing submodular in the active robot set, and (without loss of generality) $\metric$ is non-negative and $\metric[u_{1:T}(\emptyset)]=0$, then, it is: 
\begin{equation}\label{ineq:bound_cd_1}
\frac{J(u^{cd}_{1:T}(\calV))}{J(u^*_{1:T}(\calV))} \geq \frac{1}{2}.
\end{equation}
    
\item If the objective function $\metric$ is non-decreasing in the active robot set, and (without loss of generality) $\metric$ is non-negative and $\metric[u_{1:T}(\emptyset)]=0$, then, it is: 
\begin{equation}\label{ineq:bound_cd_2}
\frac{J(u^{cd}_{1:T}(\calV))}{J(u^*_{1:T}(\calV))}\geq \frac{1-c_\metric}{2}.
\end{equation}
\end{itemize}
\end{mylemma}

\paragraph*{Proof of Lemma~\ref{lem:generalized_cd}} For notational simplicity, assume an ordering among the robots in $\calV$, and let $\calV=\{1,2,\ldots,n\}$, and  $u_{\calA} \triangleq u_{1:T}(\calA)$ for some set $\calA$ of active robots. Moreover, let $J(u^{a}_{\calA},u^{b}_{\calB})$ be the value of the objective function when the robots in set $\calA$ design controls with a scheme $a$, and robots in set $\calB$ design controls with scheme $b$. Then:
\begin{itemize}
\item if the objective function $\metric$ is non-decreasing and submodular in the active robot set, and (without loss of generality) $\metric$ is non-negative and $\metric[u_{1:T}(\emptyset)]=0$, then:
\begin{alignat}{3}
\label{cd:ineq_mono}J(u^*_{1:n}) &\leq J(u^*_{1:n}) + \sum_{i=1}^n [J(u^{cd}_{1:i}, u^*_{i+1:n}) \\ 
&\hspace{25mm}-J(u^{cd}_{1:i-1}, u^*_{i+1:n})] \nonumber \\
&\label{cd:eq_arrange} = J(u^{cd}_{1:n}) + \sum_{i=1}^{n}[ J(u^{cd}_{1:i-1}, u^*_{i:n})  \\
&\hspace{25mm} -J(u^{cd}_{1:i-1}, u^*_{i+1:n})]\nonumber \\
 &= J(u^{cd}_{1:n}) + \sum_{i=1}^n J(u^*_i | \{u^{cd}_{1:i-1}, u^*_{i+1,n} \})\label{cd:eq_marginal} \\
&\leq J(u^{cd}_{1:n}) + \sum_{i=1}^n J(u^*_i | u^{cd}_{1:i-1}) \label{cd:ineq_submod}\\
&\leq J(u^{cd}_{1:n}) + \sum_{i=1}^n J(u^{cd}_i | u^{cd}_{1:i-1})\label{cd:ineq_cd} 
\end{alignat}

\begin{alignat}{3}
&= J(u^{cd}_{1:n}) + J(u^{cd}_{1:n}) \label{cd:eq_marginal_sum} \\
&\leq 2J(u^{cd}_{1:n}), \label{cd:eq_marginal_sum_final}
\end{alignat}
where ineq.~(\ref{cd:ineq_mono}) holds due to monotonicity of $J$; eq.~\ref{cd:eq_arrange}) is a shift in indexes of the first term in the sum; eq.~(\ref{cd:eq_marginal}) is an expression of the sum as a sum of marginal gains; ineq.~(\ref{cd:ineq_submod}) holds due to submodularity; ineq.~(\ref{cd:ineq_cd}) holds by the coordinate-descent policy (per~eq.~\eqref{cd_policy}); eq.~(\ref{cd:eq_marginal_sum}) holds due to the definition of the marginal gain symbol $J(u^*_i | u^{cd}_{1:i-1})$ (for any $i=1,2,\ldots,n$) as $J(u^*_i , u^{cd}_{1:i-1})-J(u^{cd}_{1:i-1})$; finally, a re-arrangement of the terms in eq.~\eqref{cd:eq_marginal_sum_final} gives $J(u^{cd}_{1:n})/J(u^{*}_{1:n}) \geq 1/2$.

\item If the objective function $\metric$ is non-decreasing in the active robot set, and (without loss of generality) $\metric$ is non-negative and $\metric[u_{1:T}(\emptyset)]=0$, then multiplying both sides of eq.~\eqref{cd:eq_marginal} (which holds for any non-decreasing $\metric$) with $(1-c_\metric)$, we have:
\begin{alignat}{3}
 (1-&c_\metric)J(u^*_{1:n}) \nonumber\\
 &= (1-c_\metric)J(u^{cd}_{1:n}) + \nonumber\\
 &\hspace{2cm}(1-c_\metric)\sum_{i=1}^n J(u^*_i | \{u^{cd}_{1:i-1}, u^*_{i+1,n} \})\nonumber\\
 &\leq J(u^{cd}_{1:n}) + (1-c_\metric)\sum_{i=1}^n J(u^*_i | \{u^{cd}_{1:i-1}, u^*_{i+1,n} \})\label{cd2:eq_marginal2} \\
&\leq J(u^{cd}_{1:n}) + \sum_{i=1}^n J(u^*_i | u^{cd}_{1:i-1}) \label{cd2:ineq_submod}\\
&\leq J(u^{cd}_{1:n}) + \sum_{i=1}^n J(u^{cd}_i | u^{cd}_{1:i-1})\label{cd2:ineq_cd} \\
&= J(u^{cd}_{1:n}) + J(u^{cd}_{1:n}) \label{cd2:eq_marginal_sum} \\
&\leq 2J(u^{cd}_{1:n}),
\end{alignat}
where, ineq.~\eqref{cd2:eq_marginal2} holds since $0\leq c_\metric\leq 1$; ineq.~(\ref{cd2:ineq_submod}) holds since $J$ is non-decreasing in the set of active robots, and Definition~\ref{def:total_curvature} of total curvature implies that for any non-decreasing  set function $g:2^\calV\mapsto\mathbb{R}$, for any element $\elem\in\calV$, and for any set $\calA,\calB\subseteq \calV\setminus \{\elem\}$, it is: 
 \begin{equation}\label{eq:ineq_total_curvature}
 (1-c_g) g(\elem|\calB)\leq g(\{\elem\}|\calA);
  \end{equation}
ineq.~(\ref{cd2:ineq_cd}) holds by the coordinate-descent algorithm; eq.~(\ref{cd2:eq_marginal_sum}) holds due to the definition of the marginal gain symbol $J(u^*_i | u^{cd}_{1:i-1})$ (for any $i=1,2,\ldots,n$) as $J(u^*_i , u^{cd}_{1:i-1})-J(u^{cd}_{1:i-1})$; finally, a re-arrangement of terms gives $J(u^{cd}_{1:n})/J(u^{*}_{1:n}) \geq (1-c_\metric)/2$.  \hfill $\blacksquare$
\end{itemize}

\section{Proof of Theorem~\ref{th:per_alg_dec_resil_coord_decent}}

We first prove Theorem~\ref{th:per_alg_dec_resil_coord_decent}'s part 1 (approximation performance), and then, Theorem~\ref{th:per_alg_dec_resil_coord_decent}'s part 2 (communication rounds).

\subsection{Proof of Theorem~\ref{th:per_alg_dec_resil_coord_decent}'s part 1 (approximation performance)}

The proof follows the steps of the proof of~\cite[Theorem~1]{tzoumas2017resilient} and of the proof of~\cite[Theorem~1]{tzoumas2018resilientSequential}.  

\medskip 

We first prove ineq.~\eqref{ineq:bound_sub}; then, we prove ineq.~\eqref{ineq:bound_non_sub}.

To the above ends, we use the following notation (along with the notation introduced in Theorem~\ref{th:per_alg_dec_resil_coord_decent} and in Appendix~A): given that using Algorithm~\ref{alg:dec_resil_coord_decent} the robots in $\calV$ select control inputs $u_{1:T}(\calV)$, then, for notational simplicity:
\begin{itemize}
\item for any active robot set $\calR\subseteq \calV$, let $J(\calR)\triangleq J[u_{1:T}(\calR)]$.
\item let $\calA^\star\triangleq\calA^\star[u_{1:T}(\calV)]$;
\item let $\calL^+\triangleq \calL\setminus \calA^\star$, i.e., $\calS_1$ are the remaining robots in $\calL$ after the removal of the robots in $\calA^\star$;
\item let $(\calV\setminus\calL)^+\triangleq (\calV\setminus\calL)\setminus \calA^\star$, i.e., $\calS_2$ are the remaining robots in $\calV\setminus\calL$ after the removal of the robots in $\calA^\star$.
\end{itemize}

\paragraph*{Proof of ineq.~\eqref{ineq:bound_sub}}
Consider that the objective function $\metric$ is non-decreasing and submodular in the active robot set, such that (without loss of generality) $\metric$ is non-negative and $\metric[u_{1:T}(\emptyset)]=0$.  We first prove the part $1-\kappa_\metric$ of the bound in the right-hand-side of ineq.~\eqref{ineq:bound_sub}, and then, the part $h(|\calV|,\attack)$ of the bound in the right-hand-side of ineq.~\eqref{ineq:bound_sub}.

To prove the part $1-\kappa_\metric$ of the bound in the right-hand-side of ineq.~\eqref{ineq:bound_sub}, we follow the steps of the proof of~\cite[Theorem~1]{tzoumas2017resilient}, and
 make the following observations:
\begin{align}
& \metric(\mathcal{V}\setminus\mathcal{A}^\star) \nonumber\\
&\;\;\;= \metric(\calL^+\cup (\calV\setminus\calL)^+)\label{ineq:aux_14}\\
&\;\;\;\geq (1-\kappa_\metric)\sum_{v \in \calL^+\cup (\calV\setminus\calL)^+}\metric(v)\label{ineq:aux_15}\\
&\;\;\;\geq (1-\kappa_\metric)\left(\sum_{v \in (\mathcal{V}\setminus \mathcal{L})\setminus (\calV\setminus\calL)^+}\metric(v)+\sum_{v \in (\calV\setminus\calL)^+}\metric(v)\right)\label{ineq:aux_16}\\
&\;\;\;\geq (1-\kappa_\metric)\metric\{[(\mathcal{V}\setminus \mathcal{L})\setminus (\calV\setminus\calL)^+]\cup (\calV\setminus\calL)^+\}\label{ineq:aux_17}\\
&\;\;\;= (1-\kappa_\metric)\metric(\calV\setminus\calL)\label{ineq:aux_18},
\end{align}
where eq.~\eqref{ineq:aux_14} to~\eqref{ineq:aux_18} hold for the following reasons: eq.~\eqref{ineq:aux_14} follows from the definitions of the sets~$\mathcal{L}^+$ and $(\calV\setminus\calL)^+$; ineq.~\eqref{ineq:aux_15} follows from ineq.~\eqref{ineq:aux_14} due to Lemma~\ref{lem:non_total_curvature}; ineq.~\eqref{ineq:aux_16} follows from ineq.~\eqref{ineq:aux_15} because for all elements $v \in \mathcal{L}^+$ and all elements  $v' \in (\mathcal{V}\setminus \mathcal{L})\setminus (\calV\setminus\calL)^+$ it is $\metric(v)\geq \metric(v')$ (note that due to the definitions of the sets~$\mathcal{L}^+$ and $(\calV\setminus\calL)^+$ it is $|\mathcal{L}^+|=|(\mathcal{V}\setminus \mathcal{L})\setminus (\calV\setminus\calL)^+|$, that is, the number of non-removed elements in $\calL$ is equal to the number of removed elements in $\calV\setminus \calL$);  finally, ineq.~\eqref{ineq:aux_17} follows from ineq.~\eqref{ineq:aux_16} because the set function $\metric$ is submodular and, as~a result, the~submodularity Definition~\ref{def:sub} implies that for any sets $\mathcal{S}\subseteq \mathcal{V}$ and $\mathcal{S}'\subseteq \mathcal{V}$, it is  $\metric(\mathcal{S})+\metric(\mathcal{S}')\geq \metric(\mathcal{S}\cup \mathcal{S}')$~\cite[Proposition 2.1]{nemhauser78analysis}. We now complete the proof of the part $1-\kappa_\metric$ of the bound in the right-hand-side of ineq.~\eqref{ineq:bound_sub} by proving that in ineq.~\eqref{ineq:aux_18} it is:
\begin{equation}\label{ineq:main_ineq}
\metric(\calV\setminus\calL)\geq\metric^\star\!\!,
\end{equation}
when the robots in $\calV$ solve optimally the problems in Algorithm~\ref{alg:dec_resil_coord_decent}'s  step~\ref{line1:step_4}, per the statement of Theorem~\ref{th:per_alg_dec_resil_coord_decent}.  In particular, if for any active robot set $\calR\subseteq \calV$, we let $\bar{u}_{1:T}(\calR)\triangleq \{\bar{u}_{i,t}:~~\bar{u}_{i,t}\in \calU_{i,t},~~i\in\calR,~~t =1,2,\ldots,T\}$ denote a collection of control inputs to the robots in $\calR$,  then it is:
\begin{align}
\metric(\calV\setminus \calL)&\equiv\max_{\scriptsize\begin{array}{c}
\bar{u}_{i,t} \in \mathcal{U}_{i,t},i\in\calV,\\
 t=1,2\ldots,T
\end{array}}\metric[\bar{u}_{1:T}(\calV\setminus\calL)]\label{ineq:main_ineq_aux1}\\
&\geq \min_{\scriptsize\begin{array}{c}\bar{\calL}\subseteq \calV,\\
|\bar{\calL}|\leq \attack
\end{array}}\max_{\scriptsize\begin{array}{c}
\bar{u}_{i,t} \in \mathcal{U}_{i,t},i\in\calV,\\
 t=1,2\ldots,T
\end{array}}\metric[\bar{u}_{1:T}(\calV\setminus\bar{\calL})]\label{ineq:main_ineq_aux2}\\
&\geq \max_{\scriptsize\begin{array}{c}
\bar{u}_{i,t} \in \mathcal{U}_{i,t},i\in\calV,\\
 t=1,2\ldots,T
\end{array}}\min_{\scriptsize\begin{array}{c}\bar{\calL}\subseteq \calV,\\
|\bar{\calL}|\leq \attack
\end{array}}\metric[\bar{u}_{1:T}(\calV\setminus\bar{\calL})]\label{ineq:main_ineq_aux3}\\
&\equiv \metric^\star\!\!,\label{ineq:main_ineq_aux4}
\end{align}
where the ineqs.~\eqref{ineq:main_ineq_aux1}-\eqref{ineq:main_ineq_aux4} hold for the following reasons: the equivalence in eq.~\eqref{ineq:main_ineq_aux1} holds since the robots in $\calV$ solve optimally the problems in Algorithm~\ref{alg:dec_resil_coord_decent}'s  step~\ref{line1:step_4}, per the statement of Theorem~\ref{th:per_alg_dec_resil_coord_decent};~\eqref{ineq:main_ineq_aux2} holds since we minimize over the set $\calL$;~\eqref{ineq:main_ineq_aux3} holds because for any set $\hat{\calL}\subseteq \calV$ and any control inputs $\hat{u}_{1:T}(\calR)\triangleq \{\hat{u}_{i,t}:~~\hat{u}_{i,t}\in \calU_{i,t},~~i\in\calR,~~t =1,2,\ldots,T\}$:
\begin{align*}
&\max_{\scriptsize\begin{array}{c}
\bar{u}_{i,t} \in \mathcal{U}_{i,t},i\in\calV,\\
 t=1,2\ldots,T
\end{array}}\metric[\bar{u}_{1:T}(\calV\setminus\hat{\calL})]\geq \metric[\hat{u}_{1:T}(\calV\setminus\hat{\calL})] \Rightarrow\\
&\hspace{7mm}\min_{\scriptsize\begin{array}{c}\bar{\calL}\subseteq \calV,\\
|\bar{\calL}|\leq \attack
\end{array}}\max_{\scriptsize\begin{array}{c}
\bar{u}_{i,t} \in \mathcal{U}_{i,t},i\in\calV,\\
 t=1,2\ldots,T
\end{array}}\metric[\bar{u}_{1:T}(\calV\setminus\bar{\calL})]\geq \\
\vspace*{-2mm}
&\hspace{3.8cm}\min_{\scriptsize\begin{array}{c}\bar{\calL}\subseteq \calV,\\
|\bar{\calL}|\leq \attack
\end{array}}\metric[\hat{u}_{1:T}(\calV\setminus\bar{\calL})] \Rightarrow\\
\vspace{4mm}
&\hspace{6.7mm}\min_{\scriptsize\begin{array}{c}\bar{\calL}\subseteq \calV,\\
|\bar{\calL}|\leq \attack
\end{array}}\max_{\scriptsize\begin{array}{c}
\bar{u}_{i,t} \in \mathcal{U}_{i,t},i\in\calV,\\
 t=1,2\ldots,T
\end{array}}\metric[\bar{u}_{1:T}(\calV\setminus\bar{\calL})]\geq \\
&\hspace{2.7cm}\max_{\scriptsize\begin{array}{c}
\bar{u}_{i,t} \in \mathcal{U}_{i,t},i\in\calV,\\
 t=1,2\ldots,T
\end{array}}\min_{\scriptsize\begin{array}{c}\bar{\calL}\subseteq \calV,\\
|\bar{\calL}|\leq \attack
\end{array}}\metric[\bar{u}_{1:T}(\calV\setminus\bar{\calL})],
\end{align*}
where the last one is eq.~\eqref{ineq:main_ineq_aux3}; 
finally, the equivalence in eq.~\eqref{ineq:main_ineq_aux4} holds since $\metric^\star$ (per the statement of Theorem~\ref{th:per_alg_dec_resil_coord_decent}) denotes the optimal value to Problem~\ref{pr:resil_active_inf_acq}. 
Overall, we proved that ineq.~\eqref{ineq:main_ineq_aux4} proves ineq.~\eqref{ineq:main_ineq}; and, now, the combination of ineq.~\eqref{ineq:aux_18} and ineq.~\eqref{ineq:main_ineq} proves the part $1-\kappa_\metric$ of the bound in the right-hand-side of ineq.~\eqref{ineq:bound_sub}.

 \begin{figure}[t]
\def \setAone{ (0.25,0) circle (1cm) }
\def \setBone{ (.75,0) circle (0.4cm)}
\def \setAtwo{ (2.75,0) circle (1cm) }
\def \setBtwo{ (2.9,0) circle (0.4cm)}
\def \myrectangle{ (-1.5, -1.5) rectangle (4, 1.5) }
\begin{center}
\begin{tikzpicture}
\draw \myrectangle node[below left]{$\mathcal{V}$};
\draw \setAone node[left]{$\mathcal{L}$};
\draw \setBone node[]{$\mathcal{A}_1^\star$};
\draw \setBtwo node[]{$\mathcal{A}_2^\star$};
\end{tikzpicture}
\end{center}
\caption{\small Venn diagram, where the set $\mathcal{L}$ is the robot set defined in step~\ref{line1:step_2} of Algorithm~\ref{alg:dec_resil_coord_decent}, and the set  $\mathcal{A}_1^\star$ and the set $\mathcal{A}_2^\star$ are such that  $\mathcal{A}_1^\star=\mathcal{A}^\star\cap\mathcal{L}$, and $\mathcal{A}_2^\star=\mathcal{A}^\star\cap(\calV\setminus\mathcal{L})$ (observe that these definitions imply $\mathcal{A}_1^\star\cap \mathcal{A}_2^\star=\emptyset$ and $\mathcal{A}^\star=\mathcal{A}_1^\star\cup \mathcal{A}_2^\star$).  
}\label{fig:venn_diagram_for_proof}
\end{figure}
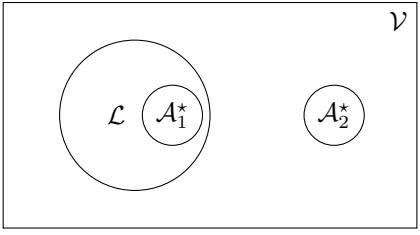

We finally prove the part $1/(1+\alpha)$ of the bound in the right-hand-side of ineq.~\eqref{ineq:bound_sub}, and complete this way the proof of Theorem~\ref{th:per_alg_dec_resil_coord_decent}.  
To this end, we follow the steps of the proof of~\cite[Theorem~1]{tzoumas2017resilient}, and use the notation introduced in Fig.~\ref{fig:venn_diagram_for_proof}, along with the following notation:
\begin{align}
	\eta \triangleq\frac{\metric(\mathcal{A}_2^\star|\calV\setminus \calA^\star)}{\metric(\mathcal{V}\setminus \calL)}
\end{align}
Later in this proof, we prove $0\leq \eta\leq 1$.  We first observe that:
\begin{equation}\label{ineq:aux_1}
\metric(\calV\setminus \calA^\star)\geq\max\{\metric(\calV\setminus \calA^\star),\metric(\calL^+)\};
\end{equation}
in the following paragraphs, we prove the three inequalities:
\begin{align}
\metric(\calV\setminus \calA^\star)&\geq(1-\eta)\metric(\mathcal{V}\setminus \calL)\label{ineq:aux_2},\\
\metric(\calL^+)&\geq \eta \frac{1}{\attack}\metric(\mathcal{V}\setminus \calL),\label{ineq:aux_3}\\
\max\{(1-\eta),\eta\frac{1}{\attack}\}&\geq \frac{1}{\attack+1}.\label{ineq:aux_4}
\end{align}
Then, if we substitute ineq.~\eqref{ineq:aux_2}, ineq.~\eqref{ineq:aux_3} and ineq.~\eqref{ineq:aux_4} to ineq.~\eqref{ineq:aux_1}, and take into account that $\metric(\mathcal{V}\setminus \calL)\geq 0$, then:
\begin{equation*}
\metric(\calV\setminus \calA^\star)\geq \frac{1}{\attack+1}\metric(\mathcal{V}\setminus \calL),
\end{equation*}
which implies the part $1/(1+\alpha)$ of the bound in the right-hand-side of ineq.~\eqref{ineq:bound_sub}, after taking into account ineq.~\eqref{ineq:main_ineq}.

We next complete the proof of the part $1/(1+\alpha)$ of the bound in the right-hand-side of ineq.~\eqref{ineq:bound_sub} by proving $0\leq \eta\leq 1$, ineq.~\eqref{ineq:aux_2}, ineq.~\eqref{ineq:aux_3}, and ineq.~\eqref{ineq:aux_4}.

\setcounter{paragraph}{0}
\paragraph{Proof of ineq.~$0\leq \eta\leq 1$} We first prove $\eta\geq 0$, and then $\eta\leq 1$:~$\eta\geq 0$, since $\eta\equiv\metric(\mathcal{A}_2^\star|\calV\setminus \calA^\star)/\metric(\mathcal{V}\setminus\calL)$, and $\metric$ is non-negative; and~$\eta\leq 1$, since $\metric(\mathcal{V}\setminus\calL)\geq \metric(\mathcal{A}^\star_2)$, due to monotonicity of $\metric$ and that $\mathcal{A}^\star_2 \subseteq \mathcal{V}\setminus\calL$, and $\metric(\mathcal{A}^\star_2)\geq \metric(\mathcal{A}_2^\star|\calV\setminus \calA^\star)$, due to submodularity of $\metric$ and that $\emptyset \subseteq \calV\setminus \calA^\star$.

\paragraph{Proof of ineq.~\eqref{ineq:aux_2}}  We complete the proof of ineq.~\eqref{ineq:aux_2} in two steps.  First, it can be verified that:
\begin{align}\label{eq:aux_1}
& f(\calV\setminus \calA^\star)=f(\mathcal{V}\setminus\calL)-\nonumber\\ & \metric(\mathcal{A}^\star_2|\calV\setminus \calA^\star)+\metric(\mathcal{L}|\mathcal{V}\setminus\calL)-\metric(\mathcal{A}^\star_1|\mathcal{V}\setminus\mathcal{A}^\star_1),
\end{align}
since for any sets $\mathcal{X}\subseteq \mathcal{V}$ and $\mathcal{Y}\subseteq \mathcal{V}$, it is $\metric(\mathcal{X}|\mathcal{Y})\equiv\metric(\mathcal{X}\cup \mathcal{Y})-\metric(\mathcal{Y})$. Second, eq.~\eqref{eq:aux_1} implies ineq.~\eqref{ineq:aux_2}, since $\metric(\mathcal{A}^\star_2|\calV\setminus \calA^\star)=\eta \metric(\mathcal{V}\setminus\calL)$, and $\metric(\mathcal{L}|\mathcal{V}\setminus\calL)-\metric(\mathcal{A}^\star_1|\mathcal{V}\setminus\mathcal{A}^\star_1)\geq 0$;
the latter is true due to the following two observations:~$\metric(\mathcal{L}|\mathcal{V}\setminus\calL)\geq \metric(\mathcal{A}_1^\star|\mathcal{V}\setminus\calL)$, since $\metric$ is monotone and $\mathcal{A}_1^\star \subseteq \mathcal{L}$; and~$\metric(\mathcal{A}_1^\star|\mathcal{V}\setminus\calL)\geq \metric(\mathcal{A}^\star_1|\mathcal{V}\setminus\mathcal{A}^\star_1)$, since $\metric$ is submodular and $\mathcal{V}\setminus\calL\subseteq \mathcal{V}\setminus\mathcal{A}^\star_1$ (see also Fig.~\ref{fig:venn_diagram_for_proof}).

\paragraph{Proof of ineq.~\eqref{ineq:aux_3}} 
To prove ineq.~\eqref{ineq:aux_3}, since it is  $\calA^\star_2\neq \emptyset$ (and, as a result, also $\calL^+\neq \emptyset$), and for all elements $a \in \calL^+$ and all elements $b\in \calA^\star_2$, it is $\metric(a)\geq \metric(b)$,  from Lemma~\ref{lem:D3} we have:
\begin{align}
\metric(\calA^\star_2|\calL^+)&\leq |\calA^\star_2|\metric(\calL^+)\nonumber\\
&\leq \attack \metric(\calL^+),\label{aux:111}
\end{align}
since $|\calA^\star_2|\leq \attack$.  Overall,
\begin{align}
\metric(\calL^+)&\geq \frac{1}{\attack}\metric(\calA^\star_2|\calL^+)\label{aux5:1}\\
&\geq \frac{1}{\attack}\metric(\calA^\star_2|\calL^+\cup (\calV\setminus \calL)^+)\label{aux5:2}\\
\end{align}

\begin{align}
&=\frac{1}{\attack}\metric(\mathcal{A}_2^\star|\calV\setminus \calA^\star)\label{aux5:3}\\
&=\eta\frac{1}{\attack}\metric(\calV\setminus \calL),\label{aux5:4}
\end{align}
where ineq.~\eqref{aux5:1} to eq.~\eqref{aux5:4} hold for the following reasons: ineq.~\eqref{aux5:1} follows from ineq.~\eqref{aux:111}; ineq.~\eqref{aux5:2} holds since $\metric$ is submodular and $\calL^+\subseteq \calL^+\cup (\calV\setminus \calL)^+$; eq.~\eqref{aux5:3} holds due to the definitions of the sets $\calL^+$, $(\calV\setminus \calL)^+$ and $\mathcal{A}^\star$; finally, eq.~\eqref{aux5:4} holds due to the definition of $\eta$.  Overall, the~latter derivation concludes the proof of ineq.~\eqref{ineq:aux_3}.

\paragraph{Proof of ineq.~\eqref{ineq:aux_4}}  Let $b=1/\attack$.  We complete the proof first for the case where 
$(1-\eta)\geq \eta b$, and then for the case $(1-\eta)<\eta b$: i) When $(1-\eta)\geq \eta b$, $\max\{(1-\eta),\eta b\}= 1-\eta$ and $\eta \leq 1/(1+b)$.  Due to the latter, $1-\eta \geq b/(1+b)=1/(\attack+1)$ and, as a result,~\eqref{ineq:aux_4} holds. ii) When $(1-\eta)< \eta b$, $\max\{(1-\eta),\eta b\}= \eta b$ and $\eta > 1/(1+b)$. Due to the latter, $\eta b >  b/(1+b)$ and, as a result,~\eqref{ineq:aux_4} holds. 

We completed the proof of~$0\leq \eta\leq 1$, and of ineqs.~\eqref{ineq:aux_2},~\eqref{ineq:aux_3} and~\eqref{ineq:aux_4}.  Thus, we also completed the proof of the part $1/(1+\alpha)$ of the bound in the right-hand-side of ineq.~\eqref{ineq:bound_sub}, and, in sum, the proof of ineq.~\eqref{ineq:bound_sub}.

\paragraph*{Proof of ineq.~\eqref{ineq:bound_non_sub}} 
Consider that the objective function $\metric$ is non-decreasing in the active robot set, such that (without loss of generality) $\metric$ is non-negative and $\metric[u_{1:T}(\emptyset)]=0$.

The proof follows the steps of the proof of~\cite[Theorem~1]{tzoumas2018resilientSequential}, by making the following observations:
\begin{align}
& \metric(\mathcal{V}\setminus\mathcal{A}^\star) \nonumber\\
&\;\;\;= \metric(\calL^+\cup (\calV\setminus\calL)^+)\label{ineq2:aux_14}\\
&\;\;\;\geq (1-c_\metric)\sum_{v \in \calL^+\cup (\calV\setminus\calL)^+}\metric(v)\label{ineq2:aux_15}\\
&\;\;\;\geq (1-c_\metric)\left(\sum_{v \in (\mathcal{V}\setminus \mathcal{L})\setminus (\calV\setminus\calL)^+}\metric(v)+\sum_{v \in (\calV\setminus\calL)^+}\metric(v)\right)\label{ineq2:aux_16}\\
&\;\;\;\geq (1-c_\metric)^2\metric\{[(\mathcal{V}\setminus \mathcal{L})\setminus (\calV\setminus\calL)^+]\cup (\calV\setminus\calL)^+\}\label{ineq2:aux_17}\\
&\;\;\;= (1-c_\metric)^2\metric(\calV\setminus\calL)\label{ineq2:aux_18},
\end{align}
where eq.~\eqref{ineq2:aux_14} to~\eqref{ineq2:aux_18} hold for the following reasons: eq.~\eqref{ineq2:aux_14} follows from the definitions of the sets~$\mathcal{L}^+$ and $(\calV\setminus\calL)^+$; ineq.~\eqref{ineq2:aux_15} follows from ineq.~\eqref{ineq2:aux_14} due to Lemma~\ref{lem:curvature}; ineq.~\eqref{ineq2:aux_16} follows from ineq.~\eqref{ineq2:aux_15} because for all elements $v \in \mathcal{L}^+$ and all elements  $v' \in (\mathcal{V}\setminus \mathcal{L})\setminus (\calV\setminus\calL)^+$ it is $\metric(v)\geq \metric(v')$ (note that due to the definitions of the sets~$\mathcal{L}^+$ and $(\calV\setminus\calL)^+$ it is $|\mathcal{L}^+|=|(\mathcal{V}\setminus \mathcal{L})\setminus (\calV\setminus\calL)^+|$, that is, the number of non-removed elements in $\calL$ is equal to the number of removed elements in $\calV\setminus \calL$);  finally, ineq.~\eqref{ineq2:aux_17} follows from ineq.~\eqref{ineq2:aux_16} because the set function $\metric$ is non-decreasing and Corollary~\ref{cor:ineq_from_lemmata} applies.  Overall, the combination of ineq.~\eqref{ineq2:aux_18}  and ineq.~\eqref{ineq:main_ineq} (observe that ineq.~\eqref{ineq:main_ineq} still holds if the objective function $\metric$ is merely non-decreasing) proves ineq.~\eqref{ineq:bound_non_sub}. \hfill $\blacksquare$

\subsection{Proof of Theorem~\ref{th:per_alg_dec_resil_coord_decent}'s part 2 (communication rounds)}

We described the steps of Algorithm~\ref{alg:dec_resil_coord_decent} in Section~\ref{subsec:alg_description}.  In particular,
Algorithm~\ref{alg:dec_resil_coord_decent} is composed of four steps: 

\setcounter{paragraph}{0}
\paragraph{Computation of robots' marginal contributions in the absence of attacks (step~\ref{line1:step_1} of Algorithm~\ref{alg:dec_resil_coord_decent})} This step requires zero rounds of communication among the robots, since each robot $i\in \calV$, by solving the problem in eq.~\eqref{pr:marginal_gains_for_each_robot}, merely computes its own marginal contribution to the information gathering task in Problem~\ref{pr:resil_active_inf_acq} in the absence of any other robot in~$\calV\setminus\{i\}$, and in the absence of any attacks and failures.

\paragraph{Computation of robot set $\calL$ with the $\attack$ largest marginal contributions in the absence of attacks (step~\ref{line1:step_2} of Algorithm~\ref{alg:dec_resil_coord_decent})} This step requires at most $2|\calV|$ communication rounds, since in this step the robots in $\calV$ share their marginal contribution to the information gathering task, which they computed in Algorithm~\ref{alg:dec_resil_coord_decent}'s step~\ref{line1:step_1},
and decide which subset~$\calL$ of them composes a set of $\attack$ robots with the $\attack$ largest marginal contributions; this procedure can be executed with minimal communication (at most $2|\calV|$ communication rounds), e.g., by accumulating (through the communication network) to one robot all the marginal contributions $\{q_i:i\in\calV\}$, and, then, by letting this robot to select the set $\calL$, and to communicate it back to the rest of the robots. 

\paragraph{Computation of control inputs of robots in $\calL$ (step~\ref{line1:step_3} of Algorithm~\ref{alg:dec_resil_coord_decent})} This steps requires zero rounds of communication among the robots, since each robot in the set $\calL$, per Algorithm~\ref{alg:dec_resil_coord_decent}'s step~\ref{line1:step_2}, merely adopts the controls it computed in Algorithm~\ref{alg:dec_resil_coord_decent}'s step~\ref{line1:step_1} (e.g., using the algorithm in~\cite{atanasov2014information}).

\paragraph{Computation of control inputs of robots in $\calV\setminus \calL$ (step~\ref{line1:step_4} of Algorithm~\ref{alg:dec_resil_coord_decent})} This step is executed in $\rho$ rounds per the statement of Theorem~\ref{th:per_alg_dec_resil_coord_decent}.

\medskip

In sum, Algorithm~\ref{alg:dec_resil_coord_decent} requires $2|\calV|+\rho$ rounds of communication among the robots in $\calV$ to terminate.  \hfill $\blacksquare$
 
\section{Proof of Proposition~\ref{prop:alg_per_with_coordinate_descent}}

We first prove Proposition~\ref{prop:alg_per_with_coordinate_descent}'s part 1 (approx.~bounds), and then, Proposition~\ref{prop:alg_per_with_coordinate_descent}'s part 2 (communication rounds).

\subsection{Proof of Proposition~\ref{prop:alg_per_with_coordinate_descent}'s part 1 (approximation bounds)}

The proof follows the steps of the proof of~Theorem~\ref{th:per_alg_dec_resil_coord_decent}; hence, we describe here only the steps where the proof differs.

\medskip

We first prove ineq.~\eqref{ineq:bound_sub_cd}; then, we prove ineq.~\eqref{ineq:bound_non_sub_cd}.

\paragraph*{Proof of ineq.~\eqref{ineq:bound_sub_cd}}
% The proof follows the steps of the proof of~\cite[Theorem~1]{tzoumas2017resilient}.
Consider that the objective function $\metric$ is non-decreasing and submodular in the active robot set, such that (without loss of generality) $\metric$ is non-negative and $\metric[u_{1:T}(\emptyset)]=0$.
Since, per Proposition~\ref{prop:alg_per_with_coordinate_descent}, Algorithm~\ref{alg:dec_resil_coord_decent} calls the coordinate descent algorithm in step~4,  the equivalence in eq.~\eqref{ineq:main_ineq_aux1} is now invalid, and, in particular, using Lemma~\ref{lem:generalized_cd}, the following inequality holds instead:
\begin{equation}\label{ineq:opt_guar_aux}
\metric(\calV\setminus \calL)\geq\frac{1}{2}\max_{\scriptsize\begin{array}{c}
\bar{u}_{i,t} \in \mathcal{U}_{i,t},i\in\calV,\\
 t=1,2\ldots,T
\end{array}}\metric[\bar{u}_{1:T}(\calV\setminus\calL)].
\end{equation}
Using ineq.~\eqref{ineq:opt_guar_aux}, and following the same steps as in eqs.~\eqref{ineq:main_ineq_aux1}-\eqref{ineq:main_ineq_aux4}, we conclude:
\begin{equation}
\metric(\calV\setminus \calL)\geq \frac{1}{2} \metric^\star\!\!.\label{ineq4:main_ineq_aux4}
\end{equation}
Using ineq.~\eqref{ineq4:main_ineq_aux4} the same way that ineq.~\eqref{ineq:main_ineq} was used in the proof of Theorem~\ref{th:per_alg_dec_resil_coord_decent}'s part 1, ineq.~\eqref{ineq:bound_non_sub_cd} is proved. 

\paragraph*{Proof of ineq.~\eqref{ineq:bound_non_sub_cd}} 
Consider that the objective function $\metric$ is non-decreasing in the active robot set, such that (without loss of generality) $\metric$ is non-negative and $\metric[u_{1:T}(\emptyset)]=0$. Similarly with the observations we made in the proof of ineq.~\eqref{ineq:bound_sub_cd}, since, per Proposition~\ref{prop:alg_per_with_coordinate_descent}, Algorithm~\ref{alg:dec_resil_coord_decent} calls the coordinate descent algorithm in step~4,  the equivalence in eq.~\eqref{ineq:main_ineq_aux1} is now invalid, and, in particular, using Lemma~\ref{lem:generalized_cd}, the following inequality holds instead:
\begin{equation}\label{ineq:opt_guar_aux_total}
\metric(\calV\setminus \calL)\geq\frac{1-c_\metric}{2}\max_{\scriptsize\begin{array}{c}
\bar{u}_{i,t} \in \mathcal{U}_{i,t},i\in\calV,\\
 t=1,2\ldots,T
\end{array}}\metric[\bar{u}_{1:T}(\calV\setminus\calL)].
\end{equation}
Using ineq.~\eqref{ineq:opt_guar_aux_total}, and following the same steps as in eqs.~\eqref{ineq:main_ineq_aux1}-\eqref{ineq:main_ineq_aux4}, we conclude:
\begin{equation}
\metric(\calV\setminus \calL)\geq \frac{1-c_\metric}{2} \metric^\star\!\!.\label{ineq5:main_ineq_aux4}
\end{equation}
Using ineq.~\eqref{ineq5:main_ineq_aux4} the same way that ineq.~\eqref{ineq:main_ineq} was used in the proof of Theorem~\ref{th:per_alg_dec_resil_coord_decent}'s part 1, ineq.~\eqref{ineq:bound_non_sub_cd} is proved. 
\hfill $\blacksquare$

\subsection{Proof of Proposition~\ref{prop:alg_per_with_coordinate_descent}'s part 2 (communication rounds)}

The description of the generalized coordinate descent in Appendix~\ref{app:prelim} implies that the generalized coordinate descent terminates in at most $|\calV|$ rounds, since each robot in $\calV$ needs to communicate with at most one robot in $\calV$ and at most once.  Therefore, per  the notation in Theorem~\ref{th:per_alg_dec_resil_coord_decent}, for the generalized coordinate descent it is $\rho=|\calV|$. Overall, per Theorem~\ref{th:per_alg_dec_resil_coord_decent}'s part~2, when Algorithm~\ref{alg:dec_resil_coord_decent} calls generalized coordinate descent in step~4, it requires $2|\calV|+\rho=3|\calV|$ rounds of communication among the robots in $\calV$ to terminate. \hfill $\blacksquare$

\end{document}